\documentclass[10pt,journal,compsoc]{IEEEtran}

\ifCLASSOPTIONcompsoc
  \usepackage[nocompress]{cite}
\else
  \usepackage{cite}
\fi

\ifCLASSINFOpdf
   \usepackage[pdftex]{graphicx}
\else
\fi

\usepackage{booktabs}% ！！
\usepackage{subfigure}
\usepackage{graphicx}
\usepackage{amsfonts}
\usepackage{algorithm}
\usepackage{algorithmic}
\usepackage{enumerate}
\usepackage{amsmath}
\usepackage{mathrsfs}
\usepackage[square, comma, sort&compress, numbers]{natbib}
\usepackage{bbm}
\usepackage{multirow}
\begin{document}
%
% paper title
% Titles are generally capitalized except for words such as a, an, and, as,
% at, but, by, for, in, nor, of, on, or, the, to and up, which are usually
% not capitalized unless they are the first or last word of the title.
% Linebreaks \\ can be used within to get better formatting as desired.
% Do not put math or special symbols in the title.
\title{Intrinsic Weight Learning Approach for Multi-view Clustering}
%
%
% author names and IEEE memberships
% note positions of commas and nonbreaking spaces ( ~ ) LaTeX will not break
% a structure at a ~ so this keeps an author's name from being broken across
% two lines.
% use \thanks{} to gain access to the first footnote area
% a separate \thanks must be used for each paragraph as LaTeX2e's \thanks
% was not built to handle multiple paragraphs
%
%
%\IEEEcompsocitemizethanks is a special \thanks that produces the bulleted
% lists the Computer Society journals use for "first footnote" author
% affiliations. Use \IEEEcompsocthanksitem which works much like \item
% for each affiliation group. When not in compsoc mode,
% \IEEEcompsocitemizethanks becomes like \thanks and
% \IEEEcompsocthanksitem becomes a line break with idention. This
% facilitates dual compilation, although admittedly the differences in the
% desired content of \author between the different types of papers makes a
% one-size-fits-all approach a daunting prospect. For instance, compsoc
% journal papers have the author affiliations above the "Manuscript
% received ..."  text while in non-compsoc journals this is reversed. Sigh.

\author{Feiping~Nie,
        Jing~Li,
        and~Xuelong~Li,~\IEEEmembership{Fellow,~IEEE}
\IEEEcompsocitemizethanks{\IEEEcompsocthanksitem Feiping Nie and Jing Li are with the Center for Optical
Imagery Analysis and Learning, Northwestern Polytechnical University, Xi'an 710072, Shaanxi, P. R. China (email: feipingnie@gmail.com; j.lee9383@gmail.com).
\protect\hfil\break
\IEEEcompsocthanksitem Xuelong Li is with the Center for OPTical IMagery Analysis and Learning (OPTIMAL), State Key Laboratory of Transient Optics and Photonics, Xi'an Institute of Optics and Precision Mechanics, Chinese Academy of Sciences, Xi'an 710119, Shaanxi, P. R. China (email: xuelong\_li@opt.ac.cn).

% note need leading \protect in front of \\ to get a newline within \thanks as
% \\ is fragile and will error, could use \hfil\break instead.
}
}

\IEEEtitleabstractindextext{ %
\begin{abstract}
Exploiting different representations, or views, of the same object for better clustering has become very popular these days, which is conventionally called \emph{multi-view clustering}. Generally, it is essential to measure the importance of each individual view, due to some noises, or inherent capacities in description. Many previous works model the view importance as weight, which is simple but effective empirically. In this paper, instead of following the traditional thoughts, we propose a new weight learning paradigm in context of multi-view clustering in virtue of the idea of re-weighted approach, and we theoretically analyze its working mechanism. Meanwhile, as a carefully achieved example, all of the views are connected by exploring a unified Laplacian rank constrained graph, which will be a representative method to compare with other weight learning approaches in experiments. Furthermore, the proposed weight learning strategy is much suitable for multi-view data, and it can be naturally integrated with many existing clustering learners. According to the numerical experiments, the proposed intrinsic weight learning approach is proved effective and practical to use in multi-view clustering.
\end{abstract}

% Note that keywords are not normally used for peerreview papers.
 }

% make the title area
\maketitle

\IEEEdisplaynontitleabstractindextext

\IEEEpeerreviewmaketitle

\ifCLASSOPTIONcompsoc
\IEEEraisesectionheading{\section{Introduction}\label{sec:introduction}}
\else
\section{Introduction}
\fi

\IEEEPARstart{M}{any} practical applications involve data obtained from multiple sources or collected with various extractors, usually known as \emph{views}. Although each individual view can be directly used  for any specific task, they are expected to work better when appropriately combined. Therefore, how to effectively cluster such kind of data has become a very hot topic.

Straightforwardly, one can choose to fuse multiple views into a single one (e.g., concatenation) in feature space and input it to a typical clustering algorithm, such as $k$-means or its kernel version \cite{DBLP:journals/neco/ScholkopfSM98}, spectral clustering \cite{DBLP:journals/pami/ShiM00}, affinity propagation \cite{frey2007clustering}, etc. However, this approach is not physically meaningful and prone to cause overfitting in the case of a small size training sample \cite{DBLP:journals/corr/abs-1304-5634}. Multi-view clustering methods often model each particular view respectively and jointly optimize them to obtain the final clustering result. In most cases, they balance the the view efficiency and the view disagreement due to a potential assumption that multiple views are complementary as well as consistent \cite{DBLP:journals/corr/abs-1304-5634,wang2016multi}.

There are quantities of previous works that focus on clustering data with multiple representations in the last two decades. The co-training based multi-view clustering methods \cite{bickel2004multi,muthukrishnan2010edge,kumar2011co,kumar2011co-training} can be seen as the extension from the co-training classification to clustering context. The basic idea of this series works is that the views are on equal terms and they can teach each other. Therefore, the clustering results of different views may be inconsistent. For instance, even with the employed regularization, in \cite{kumar2011co}, the authors suggest randomly selecting a view when the disagreement arises. For handy multiple views, to avert the aforementioned trouble, an intuitive way is to learn a unified representation after optimization. According to the literature, this thought is extensively achieved in graph-based clustering, such as joint matrix factorization approaches \cite{tang2009clustering,xia2010multiview,liu2013multi}, graph integration \cite{cai2011heterogeneous,huang2012affinity,li2015large}, the CCA(canonical correlation analysis)-based \cite{blaschko2008correlational,chaudhuri2009multi}. These methods succeed by exploiting a shared data structure which can better encode the relationships among the instances.

Actually, how to model the view importance is essential in multi-view clustering (or even in all of multi-view learning tasks). It is due to the fact that view variance widely exists in real multi-view data. There are two common conditions: 1) some views are corrupted by noise in different degree while others are clean; 2) some views are instinctively less powerful in description than others (E.g., color moments is to HOG (Histograms of Oriented Gradients) \cite{dalal2005histograms} in the family of image features). Many works model the view importance as weight and applies it on the clustering model level, which is very simple but effective. Therefore, different weight learning strategies have been developed these years, but all of them follow the same route that the view weight is explicitly defined in the objective and then it is optimized as a target variable. However, in this paper, we propose a new weight learning approach which naturally matches the multi-view clustering and significantly improves the clustering performance. What's more, it is  usually compact and light in form. Besides applying this weight learning approach in some recent clustering learners, we present that it is easily extended to more clustering method to multi-view context. The major contributions of this work are summarized as followings:
\begin{enumerate}[1)]
\item We propose a general intrinsic weight learning approach for multi-view clustering, whose working mechanism, convergence, and time complexity are carefully discussed in the paper.
\item Apart from theoretical comparison among different weight learning strategies for multi-view clustering, we achieve them in an identical base clustering learner and then their final clustering performances and learned weights are investigated in experiments.
\item We present that the proposed weight learning approach is easily integrated into more clustering methods, whose performances significantly outperform the baselines.
\end{enumerate}

The rest of the paper is organized as follows: Section \ref{sec:related works} revisits some previous methods that handle the view variance in different ways, in which we systematically summarize several representative automatical weight learning methods. In Sections \ref{sec:IWLA} and \ref{TA}, we describe the details of intrinsic weight learning approach and the relative theoretical analysis. Then we achieve an important example in Section \ref{sec:Example}, which combines the proposed weight learning approach with a recent graph-based clustering learner. Section \ref{sec:experiments} conducts numerous experiments to evaluate the proposed approach. Finally, Section \ref{sec:conclusion} concludes this paper.
\section{Related works}
\label{sec:related works}
Many previous multi-view clustering methods have made attempts to address the view variance problem. In \cite{kumar2011co-training}, the authors points out that learning the view importance is very necessary, but they simply resort to the extra prior knowledge in the paper. Some others, such as recent approaches \cite{xia2014robust,DBLP:conf/ijcai/WangZWLFP16}, tackle this problem by modeling the separated noise in each view and learn a shared clean data structure. However, this way is not able to cover the second case of view variance because of the gap of real noises and clustering errors by a weak view. As a matter of fact, measuring the view importance with the weight is direct and useful, and many works prefer this way. For convenience, \cite{cheng2009multiview} proposes to roughly compute the weights according to the proportion of the graph volume in each view. Obviously, this strategy is manually intervening and much shallow. Practically, most works would prefer to learn the weights automatically or adaptively. That means they usually optimize the objective and weights simultaneously. In the followings, we particularly introduce several types of how the previous works learn the weights which is closely related our work.

Suppose the clustering method for $i$-th view can be reduced to the minimization of the following objective function
\begin{equation}
\label{eq2.1}
\mathop {\min }\limits_{{x_c} \in {\mathcal{C}_c},{x_s} \in {\mathcal{C}_s}} \Phi_i \left( {{x_c},{x_s}} \right),
\end{equation}
where $x_c, x_s$ denote the set of view-common variables and view-specific variables, $\mathcal{C}_c, \mathcal{C}_s$ are the proxy constraints to $x_c$ and $x_s$ respectively. Noting that each view is coupled by $x_c$, for ease of notation, we instead adopt $x$ (Formally, let $x \in \mathcal{C}_{x}$) to represent the view-common variables and ignore the view-specific variables. Given a total of $M$ available views, one can derive a weighted multi-view clustering objective by minimizing a linear combination form of
\begin{equation}
\label{eq2.2}
\mathop {\min }\limits_{x,\alpha } \sum\limits_{v = 1}^M {{\alpha _v}{\Phi _v}\left( x \right)} \quad s.t.\,\alpha  \in \mathcal{C}_{\alpha},x \in \mathcal{C}_x,
\end{equation}
where \(\alpha  = \left[ {{\alpha _1},{\alpha _2},...,{\alpha _M}} \right]\), $\mathcal{C}_{\alpha}$ denotes constraints \({\alpha _v} \ge 0,\alpha {{\bf{1}}_M} = 1\) (\({{\bf{1}}_M}\) is a $M$-dimensional column vector where each element is 1). It can be easily verified that Eq. \eqref{eq2.2} has the trivial solution: the weight of best view (which has the lowest value of objective in Eq. \eqref{eq2.1}) is assigned to 1 while others are 0s. This result is apparently contrary to the assumption that all of views are usually useful. In this perspective, the following prototypes are all designed for avoiding this over-sparse problem.
\begin{enumerate}[A.]
\item \emph{Norm Regularization (NR)}

To make the weight distribution flater, some works \cite{cai2014feature,xu2016weighted,kumar2015unsupervised,karasuyama2013multiple} add a norm regularization term, and thus the objective comes to
%\begin{equation}
%\label{eq2.3}
%\begin{split}
%&\mathop {\min }\limits_{x,\alpha } \sum\limits_{i = 1}^m {{\alpha _i}{\Phi _i}\left( x \right)} {\rm{ + }}\gamma \left\| \alpha  \right\|_2^2\quad \\
%&s.t.\,{\alpha _i} \ge 0,\alpha {{\bf{1}}_m} = 1,x \in \mathcal{C},
%\end{split}
%\end{equation}
\begin{equation}
\label{eq2.3}
\mathop {\min }\limits_{x,\alpha } \sum\limits_{v = 1}^M {{\alpha _v}{\Phi _v}\left( x \right)} {\rm{ + }}\gamma_1 \left\| \alpha  \right\|_2^2\quad \,\, s.t.\,\alpha \in \mathcal{C}_{\alpha},x \in \mathcal{C}_x,
\end{equation}
where $\gamma_1$ is a non-negative parameter which controls the degree of flatness. When $\gamma_1  \to 0$, Eq. \eqref{eq2.3} reduces to Eq. \eqref{eq2.2} and the best view will be selected. On the contrary, when $\gamma_1  \to \infty $, the equal weights will be obtained. Particularly, when $x$ is fixed, the derived subproblem is
\begin{equation}
\label{eq2.4}
\mathop {\min }\limits_\alpha  \left\| {\frac{\phi }{{2\gamma_1 }} + \alpha } \right\|_2^2\quad s.t.\,\alpha  \in {\mathcal{C}_\alpha },
\end{equation}
where \(\phi  = \left[ {{\Phi _1}\left( x \right),{\Phi _2}\left( x \right),...,{\Phi _M}\left( x \right)} \right]\).
This problem can be effectively solved by the algorithm in \cite{duchi2008efficient}, and the obtained weights are usually sparse (see the discussion therein.).
\item \emph{Entropy Regularization (ER)}.

An alternative to NR is to utilize the maximum entropy \cite{jaynes1957information} to penalize the weights. It can be described as
\begin{equation}
\mathop {\min }\limits_{x,\alpha } \sum\limits_{v = 1}^M {\big({\alpha _v}{\Phi _v}\left( x \right){\rm{ + }}\gamma_2 {\alpha _v}\log {\alpha _v}\big)} \quad s.t.\,\alpha  \in {\mathcal{C}_\alpha },x \in {\mathcal{C}_x},
\end{equation}
where $\gamma_2$ has the identical effect with $\gamma_1$ in Eq. \eqref{eq2.3}. Many previous works learn the weights in this way, such as \cite{lange2005fusion,zhang2016weighted}. Similarly, when $x$ is fixed, we give the analytical solution to the corresponding subproblem as
\begin{equation}
\label{eq2.6}
{\alpha _v} = \frac{{\exp \left( {{{\left( { - {\gamma _2} - {\Phi _v}\left( x \right)} \right)} \mathord{\left/
 {\vphantom {{\left( { - {\gamma _2} - {\Phi _v}\left( x \right)} \right)} {{\gamma _2}}}} \right.
 \kern-\nulldelimiterspace} {{\gamma _2}}}} \right)}}{{\sum\limits_{u = 1}^M {\exp \left( {{{\left( { - {\gamma _2} - {\Phi _u}\left( x \right)} \right)} \mathord{\left/
 {\vphantom {{\left( { - {\gamma _2} - {\Phi _j}\left( x \right)} \right)} {{\gamma _2}}}} \right.
 \kern-\nulldelimiterspace} {{\gamma _2}}}} \right)} }}\quad \forall 1 \le v \le M,
\end{equation}
which is also known as Gibbs distribution as in \cite{lange2005fusion}. According to Eq. \eqref{eq2.6}, it can be observed that when \({{{\Phi _v}\left( x \right)} \mathord{\left/
 {\vphantom {{{\Phi _i}\left( x \right)} {{\gamma _2}}}} \right.
 \kern-\nulldelimiterspace} {{\gamma _2}}}\) is very large, $\alpha$ will be very small. Thus, loosely speaking, this strategy also learns the sparse weights.
\item \emph{Exponent Flattening (EF)}.

Another approach to smoothen the weight distribution is to introduce a parameter as the exponent of each $\alpha_i$
\begin{equation}
\mathop {\min }\limits_{x,\alpha } \sum\limits_{v = 1}^M {\alpha _v^{{\gamma _3}}{\Phi _v}\left( x \right)} \quad s.t.\,\alpha  \in {\mathcal{C}_\alpha },x \in {\mathcal{C}_x},
\end{equation}
where \({\gamma _3} > 1\). Due to the free of the regularization term, numerous multi-view clustering works \cite{xue2015gomes,tzortzis2012kernel,xu2016weighted,zhang2016weighted,tzortzis2010multiple} have adopted it. Fixing $x$, the solution of the subproblem can be given as
\begin{equation}
\label{eq2.8}
{\alpha _v} = {{1 \mathord{\left/
 {\vphantom {1 {\sum\limits_{u = 1}^M {\left( {\frac{{{\Phi _v}\left( x \right)}}{{{\Phi _u}\left( x \right)}}} \right)} }}} \right.
 \kern-\nulldelimiterspace} {\sum\limits_{u = 1}^M {\left( {\frac{{{\Phi _v}\left( x \right)}}{{{\Phi _u}\left( x \right)}}} \right)} }}^{\frac{1}{{{\gamma _3} - 1}}}}\quad \forall 1 \le v \le M.
\end{equation}
% Affinity Aggregation for Spectral Clustering haven't figure out... Multiple Non-Redundant Spectral Clustering Views not unified representation...More graphs help?
\end{enumerate}

%Suppose in each view we utilize a function to model the clustering task and every view is coupled with the wanted unified representation.

\section{Intrinsic Weight Learning Approach}\label{sec:IWLA}
\begin{figure}[tb]
\subfigure[]{
\includegraphics[width=0.205\textwidth]{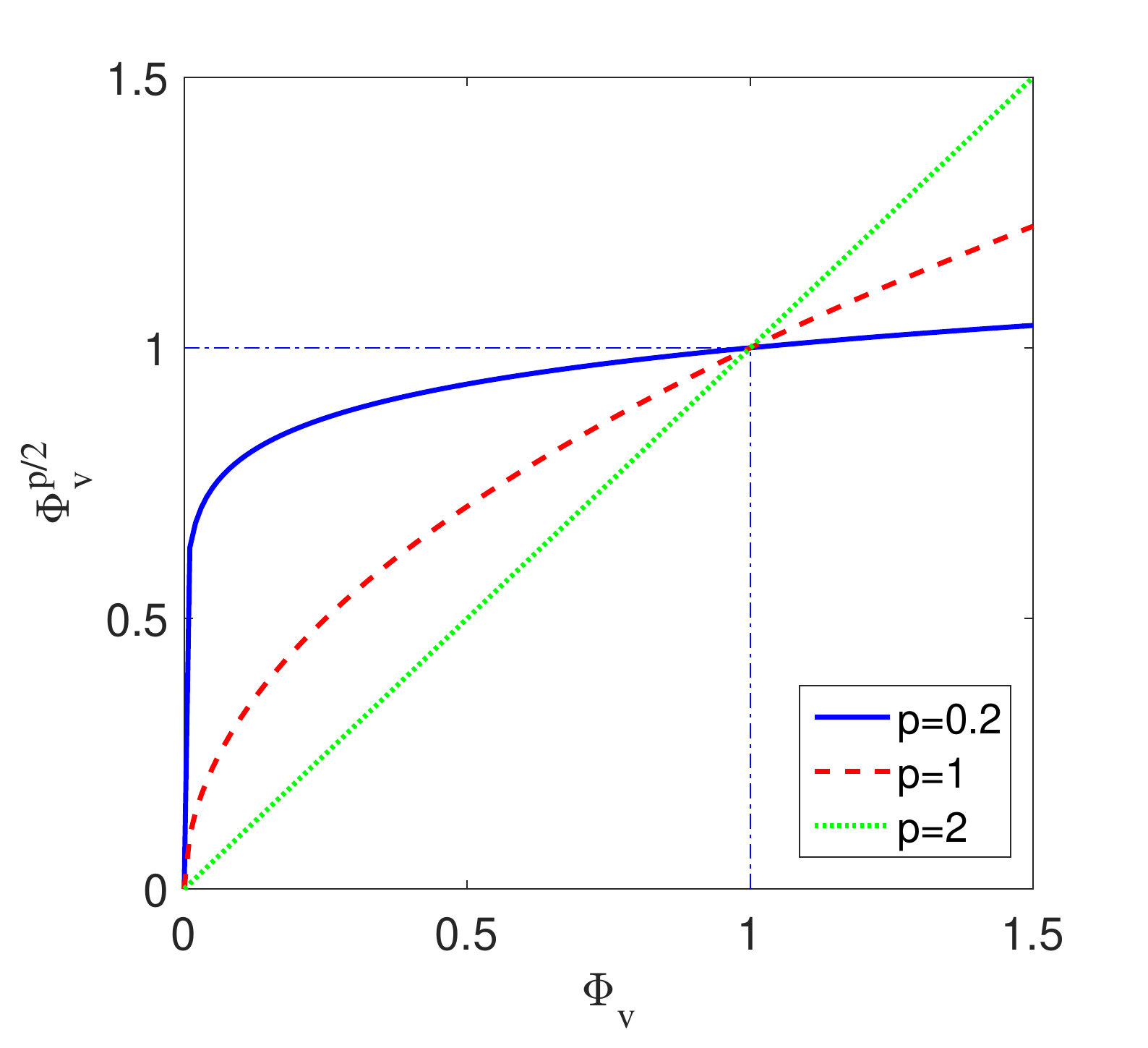}
}
\subfigure[]{
\includegraphics[width=0.22\textwidth]{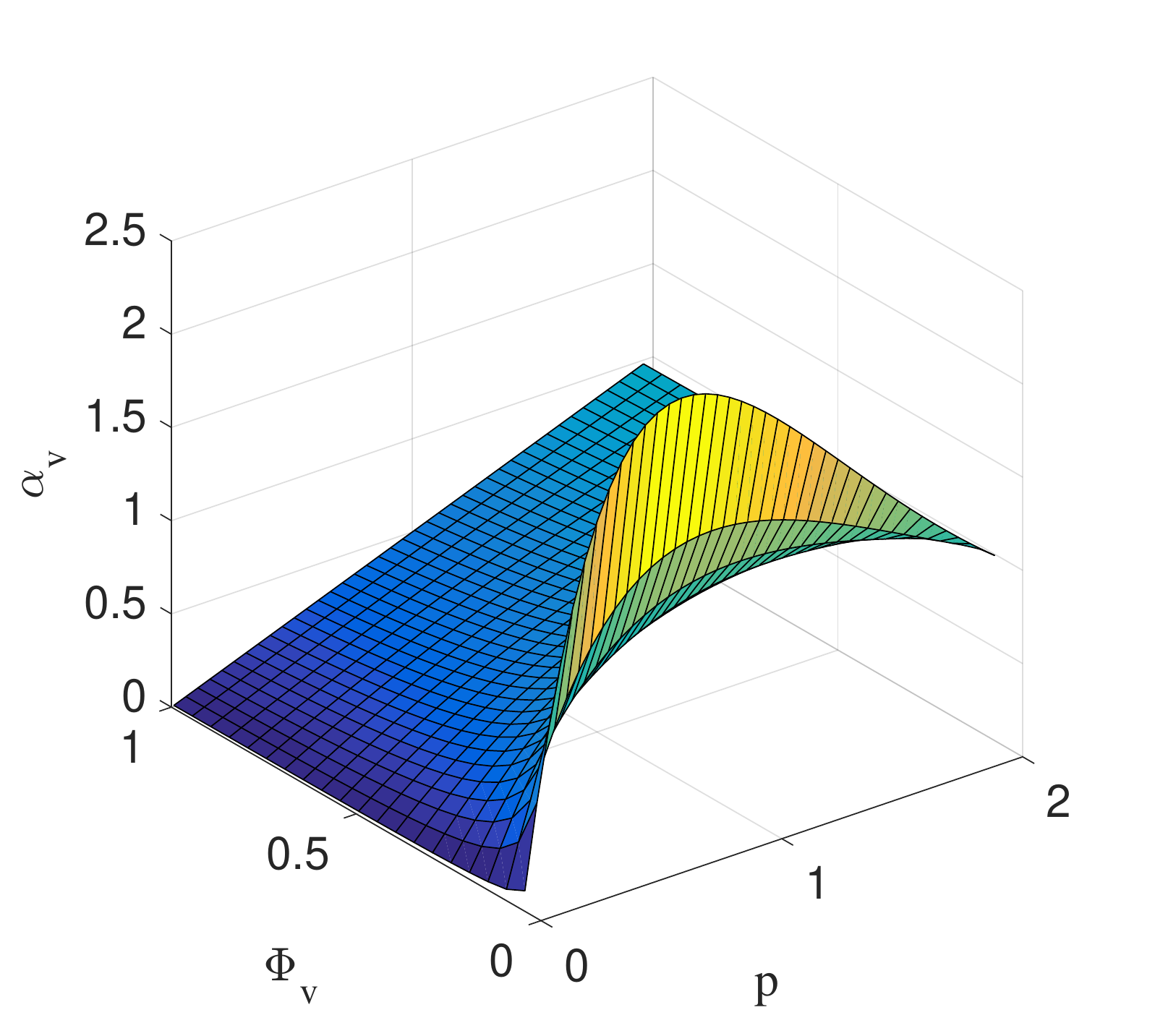}
}
\caption{(a): When $0<p \le 2$, how $\Phi_v^ {p/2}$ varies with the change of $\Phi_v$. (b) The weight $\alpha_v$ as a function of $\Phi_v$ and $p$.}
\label{f1}
\end{figure}
It is observed that previous works learn the weights by introducing a view-specific weight factor for each view in the objective, then utilizing a hyper-parameter to constrain the weight distribution, and finally solving the weights as the target variables. However, empirical studies show that they are usually sensitive to the value of the hyper-parameter (we also verify this idea in our experiments). More importantly, these methods do not touch the multi-view weights in essence because they fail to dig out the view relation according to view efficiency. To alleviate above problems, we propose a novel intrinsic weight learning approach for multi-view clustering, whose objective is
\begin{equation}
\label{eq4.1}
\mathop {\min }\limits_{x} \sum\limits_{v = 1}^M {\Phi _v^{\frac{p}{2}}\left( x \right)} \quad s.t.\,x \in {\mathcal{C}_x},
\end{equation}
where $p$ satisfies $0<p \le 2$. At the first sight, there is no weight explicitly defined in the objective, but we will see how this formulation learns intrinsic weights.

Let $\mathcal{G}(\Lambda, x)$ serve as the proxy of the constraints to $x$, and then the Lagrange function of Eq. \eqref{eq4.1} is
\begin{equation}
\label{eq4.2}
\mathop {\min }\limits_x \sum\limits_{v = 1}^M {\Phi _v^{\frac{p}{2}}(x)}  + \mathcal{G}\left(\Lambda, x \right),
\end{equation}
where $\Lambda$ represents the Lagrange multiplier. Taking the derivative of Eq. \eqref{eq4.2} w.r.t $x$ and setting the derivative to zero, we have
\begin{equation}
\label{eq4.3}
\sum\limits_{v = 1}^M {{\alpha _v}\frac{{\partial {\Phi _v}(x)}}{{\partial x}}}  + \frac{{\partial \mathcal{G}\left( {\Lambda ,x} \right)}}{{\partial x}} = 0,
\end{equation}
where
\begin{equation}
\label{eq4.4}
{\alpha _v} = \frac{p}{2}\Phi _v^{\frac{{p - 2}}{2}}\left( x \right)\quad \forall 1 \le i \le M.
\end{equation}
Since the factors of the first term in Eq. \eqref{eq4.3} are coupled with each other, Eq. \eqref{eq4.3} cannot be directly solved. However, if $\alpha_v$ is fixed, then the solution of Eq. \eqref{eq4.3} is equal to solving a linearly combined multi-view clustering problem
\begin{equation}
\label{eq4.5}
\mathop {\min }\limits_x \sum\limits_{v = 1}^M {{\alpha _v}{\Phi _v}(x)} \quad s.t.\,x \in {\mathcal{C}_x}.
\end{equation}
This problem is easier to handle. Particularly, if ${\Phi_v}(x)$ is linear w.r.t $x$, Eq. \eqref{eq4.5} will be reduced to a single clustering model but with the fused feature as the input. Then, the calculated variable $x$ can be further used to update $\alpha_v$, which inspires us to solve the problem \eqref{eq4.1} by alternatively optimizing $x$ and $\alpha_v$ iteratively. Once this procedure converges, we find that Eq. \eqref{eq4.5} is the exact form what need to be learned, and the corresponding weights are naturally obtained. In other words, solving a general problem as Eq. \eqref{eq4.1}, where the weights are not explicitly defined in the objective, actually induces a weighted linear combination of clustering models for different views. To distinguish with the aforementioned traditional weights learning methods, we call this multi-view weight learning strategy as Intrinsic Weight (IW) learning approach, which is summarized into Algorithm \ref{Alg0}.

In Fig. \ref{f1}(a), we present how the view clustering function value changes with the different powers (For ease of description, we use $\Phi_v$ as a variable when referring to $\Phi_v(x)$). When $p = 2$, Eq. \eqref{eq4.1} is equal to average weight learning. When $p$ becomes smaller, the different $\Phi_v$ will be amplified in different degrees. In one-step optimization, the weight $\alpha_v$ is computed by Eq. \eqref{eq4.4}, which is drawn into Fig. \ref{f1}(b), where $\Phi_v$ is normalized to 0-1 range. Then this approach have the following properties:

(1) \emph{For \(0 < p \le 2\), a view which has the smaller value of $\Phi_v$ will be assigned to a larger weight, and vice versa}.

For Eq. \eqref{eq4.4}, taking the partial derivative of $\alpha_v$ w.r.t $\Phi_v$, we obtain that
\[\frac{{\partial {\alpha _v}}}{{\partial {\Phi _v}}} = \frac{{p\left( {p - 2} \right)}}{4}\Phi _v^{\frac{{p - 4}}{2}} \le 0\] holds for \(0 < p \le 2\). It means that once $p$ is fixed, with the increasing of $\Phi_v$, $\alpha_v$ will be monotonically decreasing. This agrees with our knowledge, and guarantees the meaning of weights.

(2) \emph{The hyper-parameter $p$ control the smoothness of the learned weight distribution}.

Obviously, when $p = 2$, we come to the equal weights version. Now, we consider the conditions when \(0 < p < 2\). Let \({\alpha _v} = f\left( {p,{\Phi _v}} \right) = \frac{p}{2}\Phi _v^{\frac{{p - 2}}{2}}\), \({\phi _{\max }} = \max \left\{ {{\Phi _v}} \right\}_{v = 1}^M = 1\), and \({\phi _{\min }} = \min \left\{ {{\Phi _v}} \right\}_{v = 1}^M\), then the sharpest weight distribution can be approximately derived by choosing $p$ which is determined by
\[p = \mathop {\arg \max }\limits_p \left( {f\left( {p,{\phi _{\min }}} \right) - f\left( {p,{\phi _{\max }}} \right)} \right).\]
Since \(f\left( {p,1} \right) = \frac{p}{2} \in \left( {0,1} \right)\) and \({\left. {f\left( {p,{\phi _{\min }}} \right)} \right|_{{\phi _{\min }} \to 0}} = \infty \), it can be concluded that \({{\phi _{\min }}}\) plays the dominant role in choosing the hyper-parameter $p$. Taking the partial derivative of $\alpha_v$ w.r.t $p$, and setting the derivative to zero, we obtain
\[{\left. {\frac{{\partial {\alpha _v}}}{{\partial p}}} \right|_{{\Phi _v} = {\phi _{\min }}}} = {\phi _{\min }}^{\frac{{p - 2}}{2}}\left( {1 + \frac{p}{2}\ln {\phi _{\min }}} \right) = 0.\] Thus, roughly speaking, we obtain the sharpest weight distribution when \(p = \frac{{ - 2}}{{\ln {\phi _{\min }}}}\) \footnotemark.
\footnotetext{This conclusion only works in theory, because it is based on that in one-step optimization $\phi_{max} = 1$ and the range is enough to describe the smooth of a weight distribution.}

\begin{algorithm}[tb]
\caption{Intrinsic weight learning approach to solve a general problem as Eq. \eqref{eq4.1} }
\label{Alg0}
\begin{algorithmic}
\REQUIRE Hyperparameter $p$, other parameters needed for solving problem \eqref{eq2.1}.\\
Initialize the weight for each view (e.g., \(\alpha_v = \frac{1}{M}\)).
% repeat-until loop
\REPEAT
\STATE 1. Calculate $x$ by solving Eq. \eqref{eq4.5}.
\STATE 2. Update $\alpha_v$ by using Eq. \eqref{eq4.4}.
\UNTIL{converge}\\
\ENSURE $x$ used for clustering, $\alpha_v$ for each view.
\end{algorithmic}
\end{algorithm}

(3) \emph{The proposed approach learns the weights by passing the information through view-common variables.}

In each iteration in Algorithm \ref{Alg0}, given the value of $p$, it is noted from Eq. \eqref{eq4.4} that the weight $\alpha_i$ only relies on $\Phi_v$. However, according to Eq. \eqref{eq4.5}, we find  any $\Phi_j$ \((j \ne i)\) and $\alpha_i$ are correlated by the view-common variables, which is quite different from traditional weight learning approaches (see Eqs. \eqref{eq2.4}, \eqref{eq2.6}, and \eqref{eq2.8}). In other words, the aforementioned weight learning approaches are not specialized for multi-view learning, such as \cite{kumar2015unsupervised}, while this intrinsic weight learning approach is digging out the actual view relation under the view coupling assumption.

%we find that Throughout the whole procedure in Algorithm \ref{Alg0}, it is observed that weight $\alpha_i$ is computed only by $\Phi_v$.
%
%We say, this weight learning paradigm is intrinsic for multi-view learning, because the weights are acquired implicitly and are dependent on the relationship of coupled views (see Eq. \eqref{eq4.2}). Or conversely, if each view is formulated separately, the weights will be meaningless. Let us check out the aforementioned three weight learning paradigms in Section \ref{sec:related works}. An important thing is that they do not require coupling. In these paradigms, the weight of one view is directly influenced by other views. However, such weight learning are not specialized for multi-view learning, they are broad for most ensemble problems, such as multi-classifiers \cite{kumar2015unsupervised}, multi\cite{}. In this perspective, the proposed weight learning paradigm naturally suits for multi-view learning context, since views coupling assumptions are always held. Some other merits of this paradigm will be discussed according to the experiments.

%从小标题可以看出需要在上文中提炼出来新的权重学习范式
The most related works about this weight learning paradigm is re-weighed theories \cite{lawson1961contributions,beaton1974fitting,nie2014optimal}. \cite{lawson1961contributions} firstly propose the re-weight approach to solve the linear least maximum approximation problem. Later, this idea is successfully applied in compressive sensing \cite{chartrand2008iteratively}, sparse recovery \cite{daubechies2010iteratively}, and robust feature selection \cite{nie2010efficient}, etc. Different from the original re-weight theory, where the weight is employed to each single instance, \cite{nie2014optimal} extends the re-weight approach to a more general sense, i.e., utilizing the supergradients of concave functions to iteratively re-weight the concave functions. In this paper, we introduce re-weight approach into multi-view learning, but the key point stressed here is to provide a compact form to learn the view weights in multi-view clustering and demonstrate the weights efficiency.

\section{Theory Analysis}\label{TA}
This section presents the analysis of intrinsic weight learning approach in two aspects. We first prove the convergence of Algorithm \ref{Alg0}, and then analyze its time complexity.
\subsection{Convergence}
In this part, we prove that alternatively update $x$ and $\alpha_v$ in Algorithm \ref{Alg0} will monotonically decrease the objective of Eq. \eqref{eq4.1} in each iteration. First, we introduce the following lemma.

\noindent \textbf{Lemma 1:} When \(0 < p \le 2\), for any positive number $u$ and $v$, the following inequality holds:
\begin{equation}
\label{teq1}
{u^p} - \frac{p}{2}\frac{{{u^2}}}{{{v^{2 - p}}}} \le {v^p} - \frac{p}{2}\frac{{{v^2}}}{{{v^{2 - p}}}}.
\end{equation}

\noindent \textbf{Proof:} Let \(h\left( t \right) = {t^p} - \frac{p}{2}{t^2} + \frac{p}{2} - 1\), then we have.
\[{h^{'}}\left( t \right) = p{t^{p - 1}} - pt = pt\left( {{t^{p - 2}} - 1} \right).\] It is apparent that when \(t > 0\) and $0 < p \le 2$, $t = 1$ is the only zero point of \({h^{'}}\left( t \right)\). Seeing that \({h^{'}}\left( t \right) > 0 \) $(0 <t <1)$ and \({h^{'}}\left( t \right) < 0\) $\left( {t > 1} \right)$, $t =1$ is the maximum point. Since \(h(1) = 0\), when \(t > 0\) and $0<p \le 2$, $h(t)\le 0$. Therefore, let \({t^*} = {u \mathord{\left/
 {\vphantom {u v}} \right.
 \kern-\nulldelimiterspace} v}\) in $h(t)$, then \(h({u \mathord{\left/
 {\vphantom {u v}} \right.
 \kern-\nulldelimiterspace} v}) \le 0\). That is to say
\[{\left( {\frac{u}{v}} \right)^p} - \frac{p}{2}{\left( {\frac{u}{v}} \right)^2} + \frac{p}{2} - 1 \le 0.\] After a transposition, we arrive at Eq. \eqref{teq1}. \hfill $\Box$

\noindent \textbf{Theorem 1:} When $0 < p \le 2$, Algorithm \ref{Alg0} will monotonically decrease the objective in Eq. \eqref{eq4.1} in each iteration until the convergence.

\noindent \textbf{Proof:} In the $k$th iteration
\begin{equation}
\begin{split}
{x_{k + 1}} & = \mathop {\arg \min }\limits_x \sum\limits_{v = 1}^M {{\alpha _v}{\Phi _v}\left( x \right)} \\
 & = \mathop {\arg \min }\limits_x \sum\limits_{v = 1}^M {\frac{p}{2}\Phi _v^{\frac{{p - 2}}{2}}\left( {{x_k}} \right){\Phi _v}\left( x \right)},
\end{split}
\end{equation}
which means
\begin{equation}
\label{teq2}
\sum\limits_{v = 1}^M {\frac{p}{2}\Phi _v^{\frac{{p - 2}}{2}}\left( {{x_k}} \right){\Phi _v}\left( {{x_{k + 1}}} \right)}  \le \sum\limits_{v = 1}^M {\frac{p}{2}\Phi _v^{\frac{{p - 2}}{2}}\left( {{x_k}} \right){\Phi _v}\left( {{x_k}} \right)} .
\end{equation}
Let \(u = \Phi _v^{\frac{1}{2}}\left( {{x_{k + 1}}} \right)\) and \(v = \Phi _v^{\frac{1}{2}}\left( {{x_k}} \right)\), then according to Lemma 1, we obtain
\begin{equation}
\label{teq3}
\begin{split}
\sum\limits_{v = 1}^M {\left( {\Phi _v^{\frac{p}{2}}\left( {{x_{k + 1}}} \right) -  \frac{p}{2}\frac{{{\Phi _v}\left( {{x_{k + 1}}} \right)}}{{\Phi _v^{\frac{{2 - p}}{2}}\left( {{x_k}} \right)}}} \right)} \\
\le \sum\limits_{v = 1}^M {\left( {\Phi _v^{\frac{p}{2}}\left( {{x_k}} \right) - \frac{p}{2}\frac{{{\Phi _v}\left( {{x_k}} \right)}}{{\Phi _v^{\frac{{2 - p}}{2}}\left( {{x_k}} \right)}}} \right)} .
\end{split}
\end{equation}
Summing Eq. \eqref{teq2} and Eq. \eqref{teq3} in both two sides, we arrive at
\begin{equation}
\sum\limits_{v = 1}^M {\Phi _v^{\frac{p}{2}}\left( {{x_{k + 1}}} \right)}  \le \sum\limits_{v = 1}^M {\Phi _v^{\frac{p}{2}}\left( {{x_k}} \right)}.
\end{equation}
Thus, Algorithm \ref{Alg0} will monotonically decrease the objective of the problem \eqref{eq4.1} in each iteration. Obviously, since the objective must have a lower bound, the whole procedure will converge.  \hfill $\Box$
\subsection{Time Complexity}
Denote the time complexity of solving the subproblem \eqref{eq4.5} is \(\sigma \), then the time complexity of Algorithm \ref{Alg0} is \(\mathcal{O}\left( {T\left( {\sigma  + M\varepsilon } \right)} \right)\), where $\varepsilon$ is the time complexity of updating each $\alpha_v$ by Eq. \eqref{eq4.4}, and $T$ represents the number of needed iterations. Generally speaking, as \(\sigma  \gg M\varepsilon \), the total time complexity can be roughly equal to $T\sigma$. Previous work \cite{nie2014optimal} empirically showed the $T$ is no more than 50. Thus, the final time complexity is determined by the employed single clustering model.
\section{An Example CLR-IW}
\label{sec:Example}
In this section, we apply the intrinsic weight learning approach to a recent graph-based clustering method, where the specific details about the whole procedure will be presented. On the one hand, this section presents how to combine a single clustering model with the proposed weight learning strategy. One the other hand, the accomplishment helps to more accurately compare with the traditional weight learning approaches in the experiment part. %we first revisit the recent graph-based clustering method, which is the base learner in our multi-view context. Then, we introduce the a general algorithm Re-weight. Inspired by this algorithm, we propose the intrinsic weighted multi-view clustering (IWMC) method with a new weight learning strategy. Furthermore, we theoretically analyze the effectiveness.
\subsection{The Base Learner Introduction}
Given $N$ samples which can be partitioned into $C$ clusters, graph-based clustering methods usually first construct a similarity matrix to represent the affinities of all the instances. A great number of early works have studied how to design a similarity matrix with high quality, such as \cite{zelnik2005self,cai2005document}. Then, they will be input to graph-based clustering methods, e.g., spectral clustering. Finally, some postprocessings like $k$-means is employed to obtain the discrete clustering results. However, an ideal similarity matrix $S \in \mathbb{R}^{N \times N}$ is supposed to exactly have $c$ connected components, by which way, $S$ is able to be directly used for the clustering task. Recently, \cite{DBLP:conf/kdd/NieWH14,feng2014robust,DBLP:conf/uai/ChenD16,nie2016constrained} have leveraged this prpperty in different ways. We briefly introduce the Constrained Laplacian Rank (CLR) method \cite{nie2016constrained} therein, which is easier to understand and will be the base learner here.

Given an arbitrary input similarity matrix $A \in \mathbb{R}^{N \times N}$, the target similarity matrix can be learned by minimizing the following problem
\begin{equation}
\label{eq3.1.1}
\mathop {\min }\limits_{{s_i}{{\bf{1}}_N} = 1,{s_{ij}} \ge 0,S \in \mathcal{C}} {{\left\| {S - {A}} \right\|}_F^2} ,
\end{equation}
where $S$ is nonnegative, whose each row sums up to 1, and $\mathcal{C}$ represents the set of $N$ by $N$ square matrices with $C$ connected components. According to the graph theory in \cite{mohar1991laplacian,chung1997spectral}, the connectivity constraint can be replaced with a rank constraint, and thus Eq. \eqref{eq3.1.1} is specified as
\begin{equation}
\label{eq3.1.2}
\mathop {\min }\limits_{{s_i}{{\bf{1}}_N} = 1,{s_{ij}} \ge 0,rank({L_S}) = N - C} {{\left\| {S - {A}} \right\|}_F^2} ,
\end{equation}
where $rank(L_S)$ means the rank of $L_S$. The Laplacian matrix \(L_S = D_S - {{\left( {{S^T} + S} \right)} \mathord{\left/
{\vphantom {{\left( {{S^T} + S} \right)} 2}} \right.
\kern-\nulldelimiterspace} 2}\), where the degree matrix \(D_S \in {\mathbb{R}^{N \times N}}\) is defined as a diagonal matrix whose $i$-th diagonal element is \(\sum\nolimits_j {{{\left( {{s_{ij}} + {s_{ji}}} \right)} \mathord{\left/
{\vphantom {{\left( {{s_{ij}} + {s_{ji}}} \right)} 2}} \right.
\kern-\nulldelimiterspace} 2}} \). %The rank constraint to $L_S$ preserves that the target graph $S$ has exactly $c$ connected components \cite{mohar1991laplacian,chung1997spectral}, which naturally corresponds to the needed $c$ clusters.
In this way, once the target similarity matrix is solved and we can directly use it for clustering.
\subsection{The Proposed Method}
\begin{figure}[bt]
  \centering
  \includegraphics[width=0.35\textwidth]{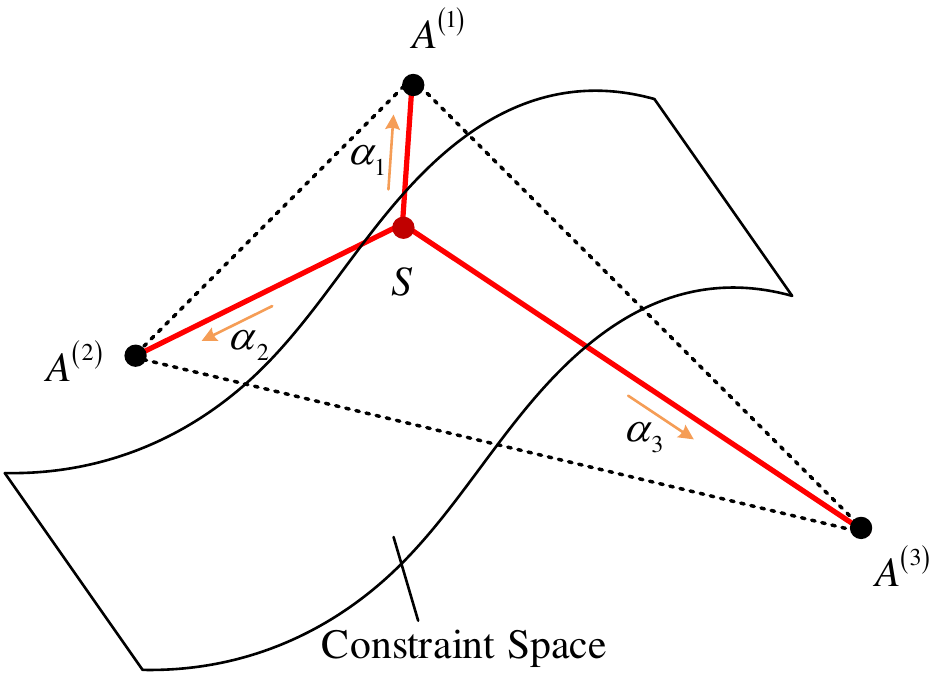}
  \caption{A three-view clustering task example. Given $A^{(1)}$, $A^{(2)}$ and $A^{(3)}$ three input views, the target similarity matrix is supposed to have the minimum of the sum of weighted distance between every input view. Simultaneously, $S$ should be guaranteed to lie in the constraint space.}\label{idea}
\end{figure}
As we stated before, it is useful to assign a weight to each view to measure its importance. Eq. \eqref{eq3.1.2} means that the target similarity matrix is expected to be as near as possible to the input one $A$. Therefore, in multi-view context, we expect to learn $S$ to be the centroid of each input $A_i$ but with the different confidence for each view. This idea is presented as in Fig. \ref{idea}. %Intuitively, one can choose any aforementioned weight learning strategy to formulate this problem, such as NR, ER, and EF. But here
We sum up each single view clustering model with the power of $p$ as indicated in Eq. \eqref{eq4.1}. The objective is written as
\begin{equation}
\label{eq3.2.1}
\mathop {\min }\limits_{{s_i}{{\bf{1}}_N} = 1,{s_{ij}} > 0,rank\left( {{L_S}} \right) = N - C} \sum\limits_{v = 1}^M {\left\| {S - {A^{\left( v \right)}}} \right\|_F^p},
\end{equation}
where \({{A^{\left( v \right)}}}\) is the $v$-th input similarity matrix, and \(0 < p \le 2\). We donate this method as CLR-IW in the following of this paper for better illustration. Following Algorithm \ref{Alg0}, we can directly write the two alternative steps of solving the problem \eqref{eq3.2.1}: solving the following linear combined CLR clustering subproblem
\begin{equation}
\label{eq3.2.2}
\mathop {\min }\limits_{{s_i}{{\bf{1}}_N} = 1,{s_{ij}} > 0,rank\left( {{L_S}} \right) = N - C} \sum\limits_{v = 1}^M {{\alpha _v}\left\| {S - {A^{\left( v \right)}}} \right\|_F^2} ,
\end{equation}
and updating the $\alpha_v$ by
\begin{equation}
\label{eq3.2.3}
{\alpha _v} = \frac{p}{2}\left\| {S - {A^{\left( v \right)}}} \right\|_F^{p-2}.
\end{equation}
It is obvious that updating $\alpha_v$ is quite simple, while solving the subproblem as Eq. \eqref{eq3.2.2} needs further calculations.

To solve Eq. \eqref{eq3.2.2}, we first let \({\rho _k}\left( {{L_S}} \right)\) to represent $k$-th smallest eigenvalue of $L_S$. Seeing that $L_S$ is positive semi-definite, \({\rho _k}\left( {{L_S}} \right) > 0\). Given a large value of $\lambda$, the rank constraint in Eq. \eqref{eq3.2.2} can be eliminated and Eq \eqref{eq3.2.2} is equal to the following form
\begin{equation}
\label{eq3.3.1}
\mathop {\min }\limits_{{s_i}{{\bf{1}}_N} = 1,{s_{ij}} > 0} \sum\limits_{v = 1}^M {{\alpha _v}\left\| {S - {A^{\left( v \right)}}} \right\|_F^2}  + 2\lambda \sum\limits_{k = 1}^C {{\rho _k}\left( {{L_S}} \right)}
\end{equation}
When $\lambda$ is large enough, note that \({\rho _k}\left( {{L_S}} \right) > 0\) for each $k$, thus the optimal solution $S$ will make \(\sum\limits_{k = 1}^C {{\rho _k}\left( {{L_S}} \right)} \) to zero and the constraint \({rank\left( {{L_S}} \right) = N - C}\) will be satisfied. Moreover, according to Ky Fan's Theory \cite{fan1950theorem}, the following equation holds
\begin{equation}
\label{eq3.3.2}
\sum\limits_{k = 1}^C {{\rho _k}\left( {{L_S}} \right)}  = \mathop {\min }\limits_{F \in {R^{N \times C}},{F^T}F = I} Tr\left( {{F^T}{L_S}F} \right).
\end{equation}
Thus, according to Eq. \eqref{eq3.3.2}, Eq. \eqref{eq3.3.1} is further written to
\begin{equation}
\label{eq3.3.3}
\begin{split}
&\mathop {\min }\limits_{S,F} \sum\limits_{v = 1}^M {{\alpha _v}\left\| {S - {A^{\left( v \right)}}} \right\|_F^2}  + 2\lambda Tr\left( {{F^T}{L_S}F} \right)\\
&s.t.\,{s_i}{{\bf{1}}_n} = 1,{s_{ij}} > 0,F \in {R^{N \times C}},{F^T}F = I.
\end{split}
\end{equation}
We solve this problem by alternatively optimizing variable $F$ and $S$ iteratively as follows.

\romannumeral1. \textbf{When $S$ is fixed}, Eq. \eqref{eq3.3.3} becomes
\begin{equation}
\label{eq3.3.5}
\mathop {\min }\limits_{F \in {\mathbb{R}^{N \times C}},{F^T}F = I} Tr\left( {{F^T}{L_S}F} \right).
\end{equation}
It is known that the optimal solution of $F$ is formed by the $C$ eigenvectors of $L_S$ corresponding to the $C$ smallest eigenvalues.

\romannumeral2. \textbf{When $F$ is fixed}, Eq. \eqref{eq3.3.3} can be written as
\begin{equation}
\label{eq3.3.6}\small
\mathop {\min }\limits_{{s_i}{{\bf{1}}_N} = 1,{s_{ij}} \ge 0} \sum\limits_{v = 1}^M {{\alpha_v}\sum\limits_{i,j = 1}^N {{{\left( {{s_{ij}} - a_{ij}^{\left( v \right)}} \right)}^2}} }  + \lambda \sum\limits_{i,j = 1}^N {\left\| {{f_i} - {f_j}} \right\|_2^2{s_{ij}}} .
\end{equation}
Since Eq. \eqref{eq3.3.6} is independent for different $i$, we turn to solve the following problem separately for each $i$:
\begin{equation}
\label{eq3.3.7}\small
\mathop {\min }\limits_{{s_i}{{\bf{1}}_N} = 1,{s_{ij}} \ge 0} \sum\limits_{j = 1}^N {\sum\limits_{v = 1}^M {{\alpha _v}{{\left( {{s_{ij}} - a_{ij}^{\left( v \right)}} \right)}^2}} }  + \lambda \sum\limits_{j = 1}^N {\left\| {{f_i} - {f_j}} \right\|_2^2{s_{ij}}} .
\end{equation}
For ease of presentation, we denote \({v_{ij}} = \left\| {{f_i} - {f_j}} \right\|_2^2\) and $v_i$ is a row vector with $j$-th element equal to $v_{ij}$ (and similarly for $s_i$ and $a_i$), Eq. \eqref{eq3.3.7} is further written in vector form as
\begin{equation}
\label{eq3.3.8}
\mathop {\min }\limits_{s_i{\bf{1}}_N = 1,{s_{ij}} \ge 0} \left\| {{s_i} - {{\left( {\sum\limits_{v = 1}^M {{\alpha^{\left( v \right)}}a_i^{\left( v \right)}}  - \frac{\lambda }{2}{v_i}} \right)} \mathord{\left/
 {\vphantom {{\left( {\sum\limits_{v = 1}^M {{w^{\left( v \right)}}a_i^{\left( v \right)}}  - \frac{\lambda }{2}{v_i}} \right)} {\sum\limits_{v = 1}^M {{w^{\left( v \right)}}} }}} \right.
 \kern-\nulldelimiterspace} {\sum\limits_{v = 1}^M {{\alpha_v}} }}} \right\|_2^2.
\end{equation}
This problem is solved just like Eq. \eqref{eq2.4}. To accelerate the computing, we can choose to update $t$ (One can set $t$ as a const, such as 10) neighbors of $i$-th data. Thus, $S$ is totally sparse and the scale of Eq. \eqref{eq3.3.8} becomes smaller. We summarize this solving process into Algorithm \ref{Alg1}.

\begin{algorithm}[tb]
\caption{The algorithm of solving Eq. \eqref{eq3.2.1} }
\label{Alg1}
\begin{algorithmic}
\REQUIRE SMs for $M$ views \(\left\{ {{A^{\left( 1 \right)}},{A^{\left( 2 \right)}},...,{A^{\left( M \right)}}} \right\}\) and \({A^{\left( v \right)}} \in {\mathbb{R}^{N \times N}}\), number of clusters $C$.\\
Initialize the weight for each view (e.g., \(\alpha_v = \frac{1}{M}\)). and Let \(S{'} = \sum\limits_{v = 1}^M {{\alpha _v}{A^{\left( v \right)}}} \).
% repeat-until loop
\REPEAT
\REPEAT
\STATE i. $S = S{'}$. Update \(F \in {R^{N \times C}}\) which is formed by the $C$ eigenvectors of $L_S$ ( \({L_S} = {D_S} - {{\left( {{S^T} + S} \right)} \mathord{\left/
 {\vphantom {{\left( {{S^T} + S} \right)} 2}} \right.
 \kern-\nulldelimiterspace} 2}\) ) corresponding to the $C$ smallest eigenvalues.
\STATE ii. Update $S$ by solving Eq. \eqref{eq3.3.8} using the algorithm proposed in \cite{duchi2008efficient}.
\UNTIL{$S$ has exactly $C$ connected compenents}

$S{'} = S$.

Update $\alpha_v$ by using Eq. \eqref{eq3.2.3}.
\UNTIL{converge}\\
\ENSURE $S \in \mathbb{R}^{N \times N}$ with exactly $C$ connected components, and the instances in each component belongs to a cluster.
\end{algorithmic}
\end{algorithm}

\section{Experiments}
\label{sec:experiments}
In this section, we firstly rely on the generated method CLR-IW to verify the effectiveness of the proposed intrinsic weight approach. Specifically, this method is compared with some primary baselines on a synthetic dataset, such as conducting CLR on each individual view, and simply assigning equal weight to every CLR model in multi-view context. Then, all of aforementioned weight learning paradigm in Section \ref{sec:related works} are horizontally compared on various multi-view datasets, where the parameter robustness and weight distribution is carefully discussed. Finally, we show that it is natural to extend the intrinsic weight learning approach to more clustering techniques.

For each method which needs to construct a graph, we use the graph building approach proposed in \cite{nie2016constrained}, since it obtains a normalized graph and only involves one parameter, the number of nearest neighbors, which is simply set as 20 in all the experiments. What's more, three standard clustering metrics, i.e., ACC \cite{cai2005document}, NMI \cite{cai2005document}, Purity \cite{varshavsky2005compact}, are used throughout all the experiments.

\subsection{Toy Examples}
\begin{figure}[tb]
\subfigure[ View 1, e = 0.6, 0.8]{
\includegraphics[width=0.23\textwidth]{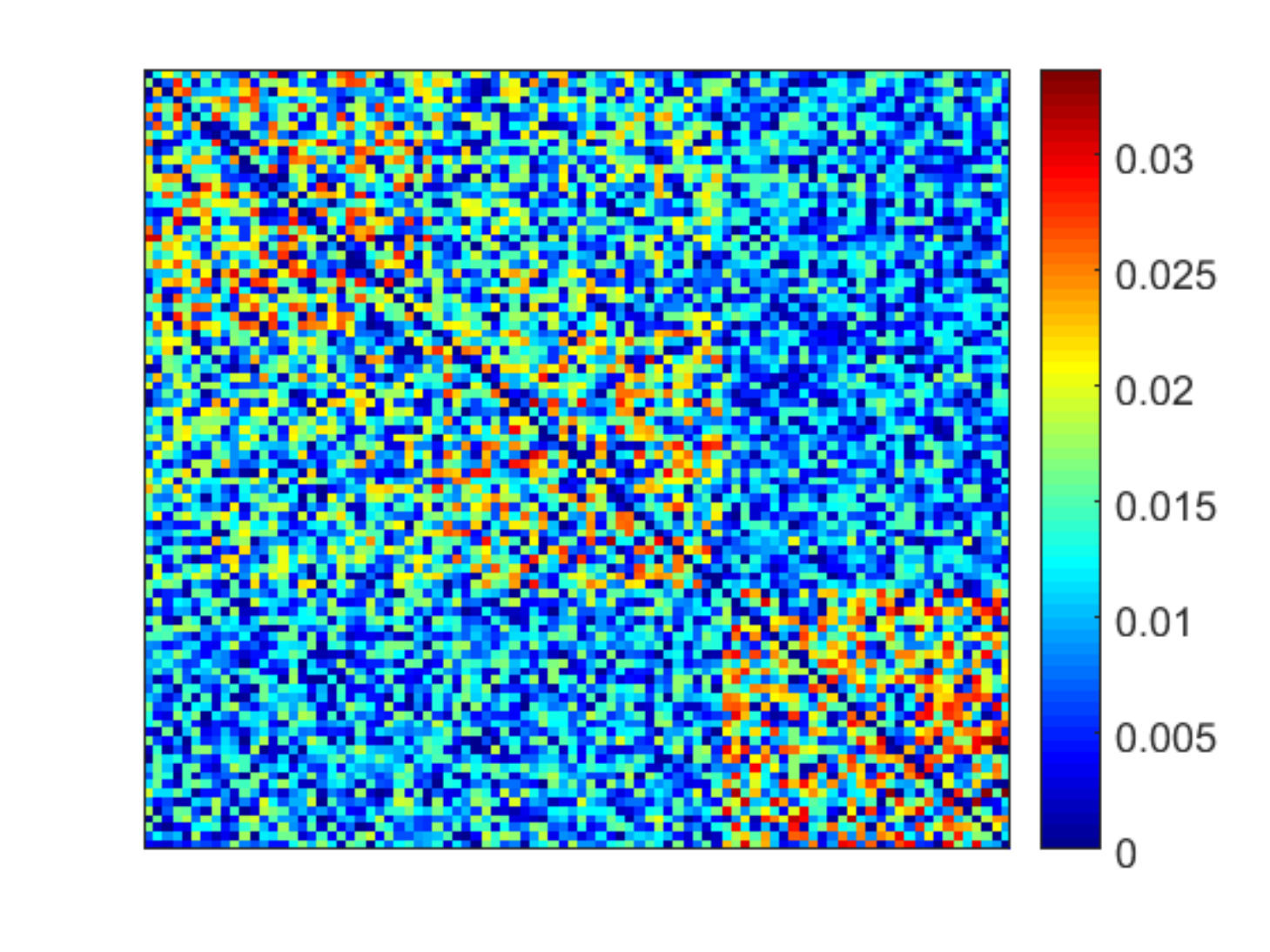}
}
\subfigure[ View 1, e = 0.6, 0.8]{
\includegraphics[width=0.23\textwidth]{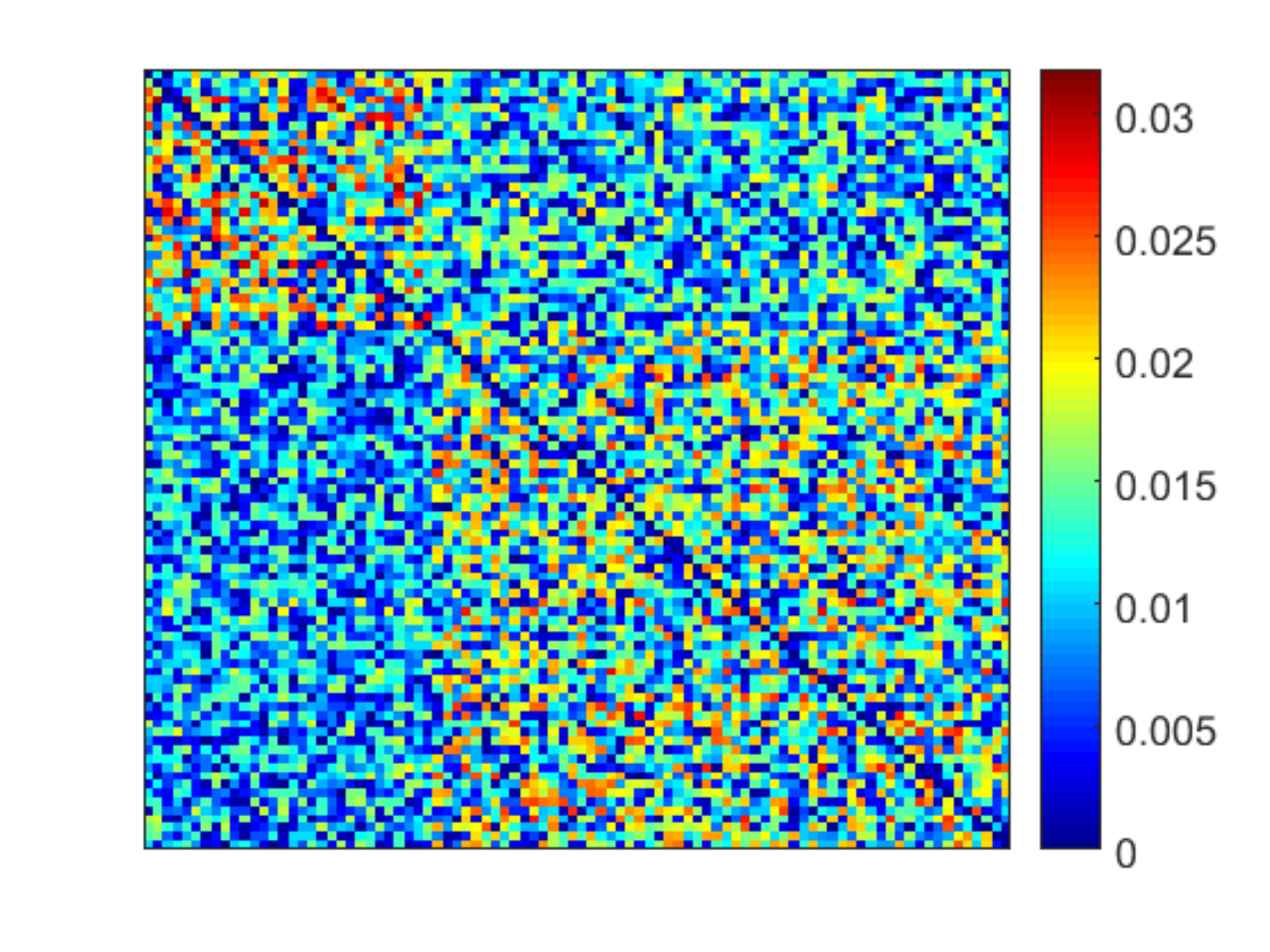}
}
\caption{This synthetic data set contains two views (a,b) which are generally complementary but with different noises.}
\label{toy1}
\end{figure}

%\begin{figure}[tb]
%\begin{minipage}{0.23\textwidth}\label{1c}
%\centerline{\includegraphics[width=4.5cm]{toy2_1.pdf}}
%\centerline{(a) View 1, e = 0.6,0.8}
%\end{minipage}
%\hfill
%\begin{minipage}{0.23\textwidth}\label{1d}
%\centerline{\includegraphics[width=4.5cm]{toy2_2.pdf}}
%\centerline{(b) View 2, e = 0.7,1.0}
%\end{minipage}
%\caption{Toy\_2 contains two views (a,b) which are generally complementary but with different noises. }
%\label{toy1}
%\end{figure}

\begin{figure*}[tb]
\subfigure[View1]{
\includegraphics[width=0.23\textwidth]{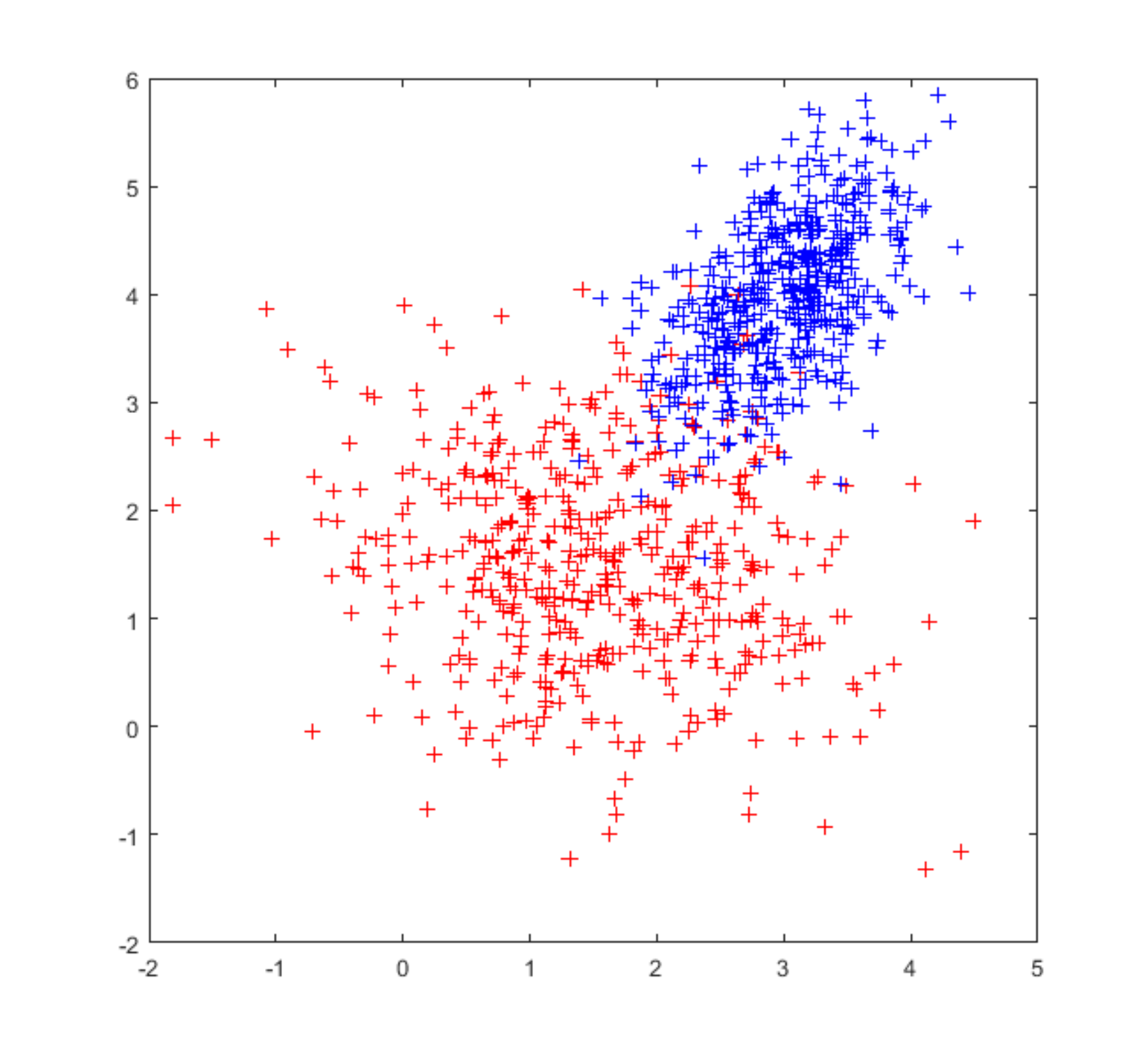}
}\label{v1}
\subfigure[$S$ learned form view 1]{
\includegraphics[width=0.23\textwidth]{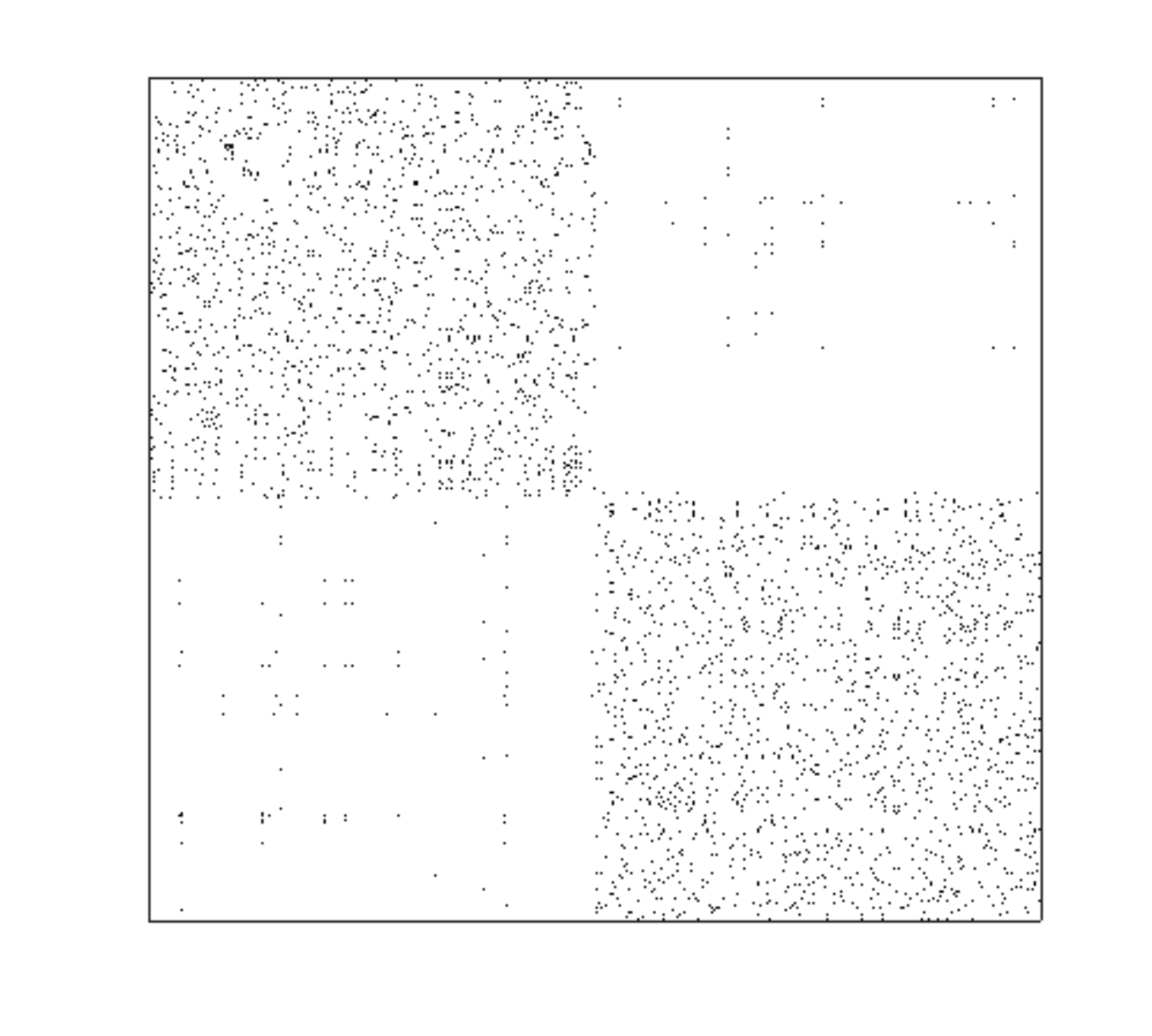}
}\label{s1}
\subfigure[View2]{
\includegraphics[width=0.23\textwidth]{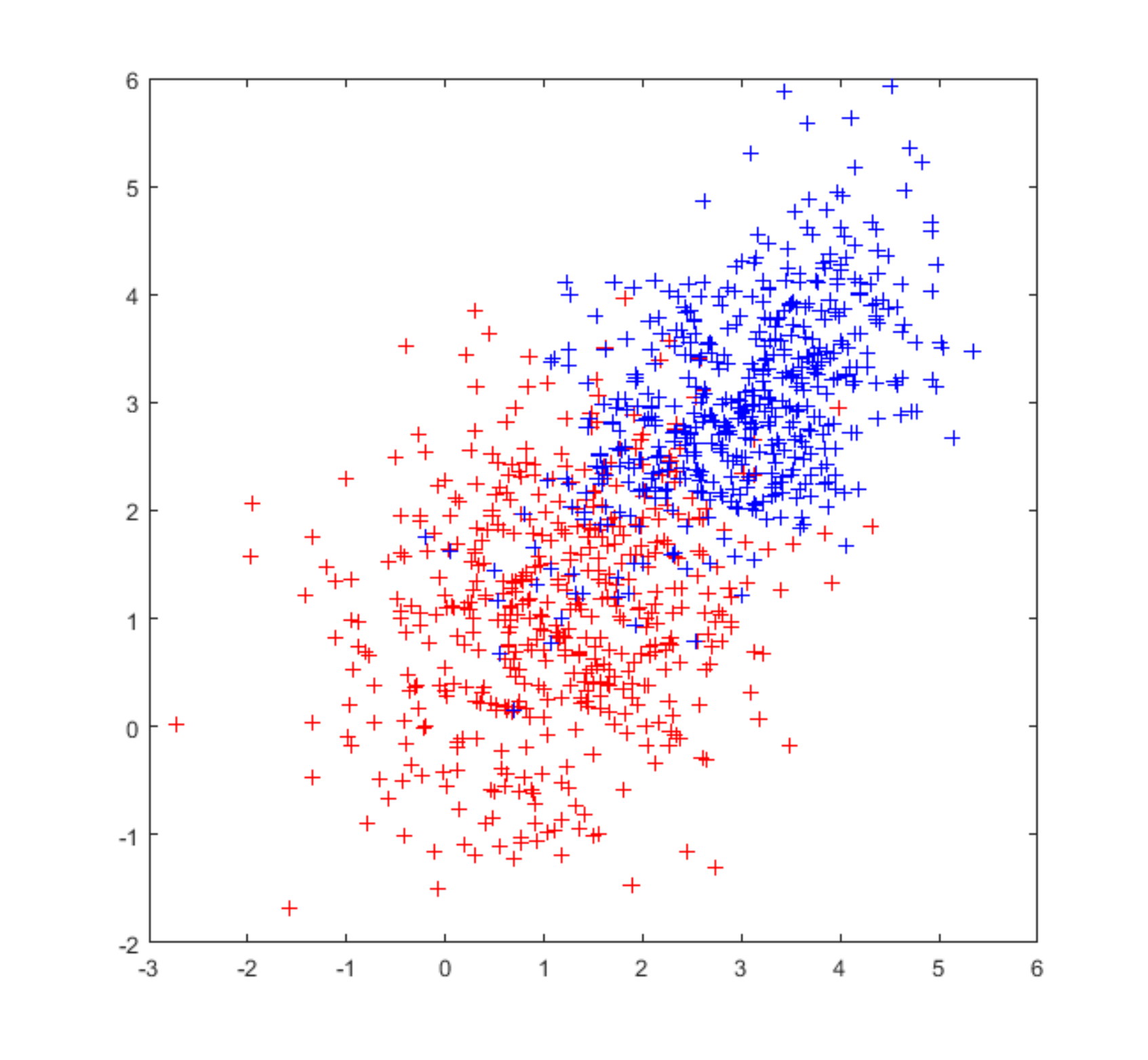}
}\label{v2}
\subfigure[$S$ learned form view 2]{
\includegraphics[width=0.23\textwidth]{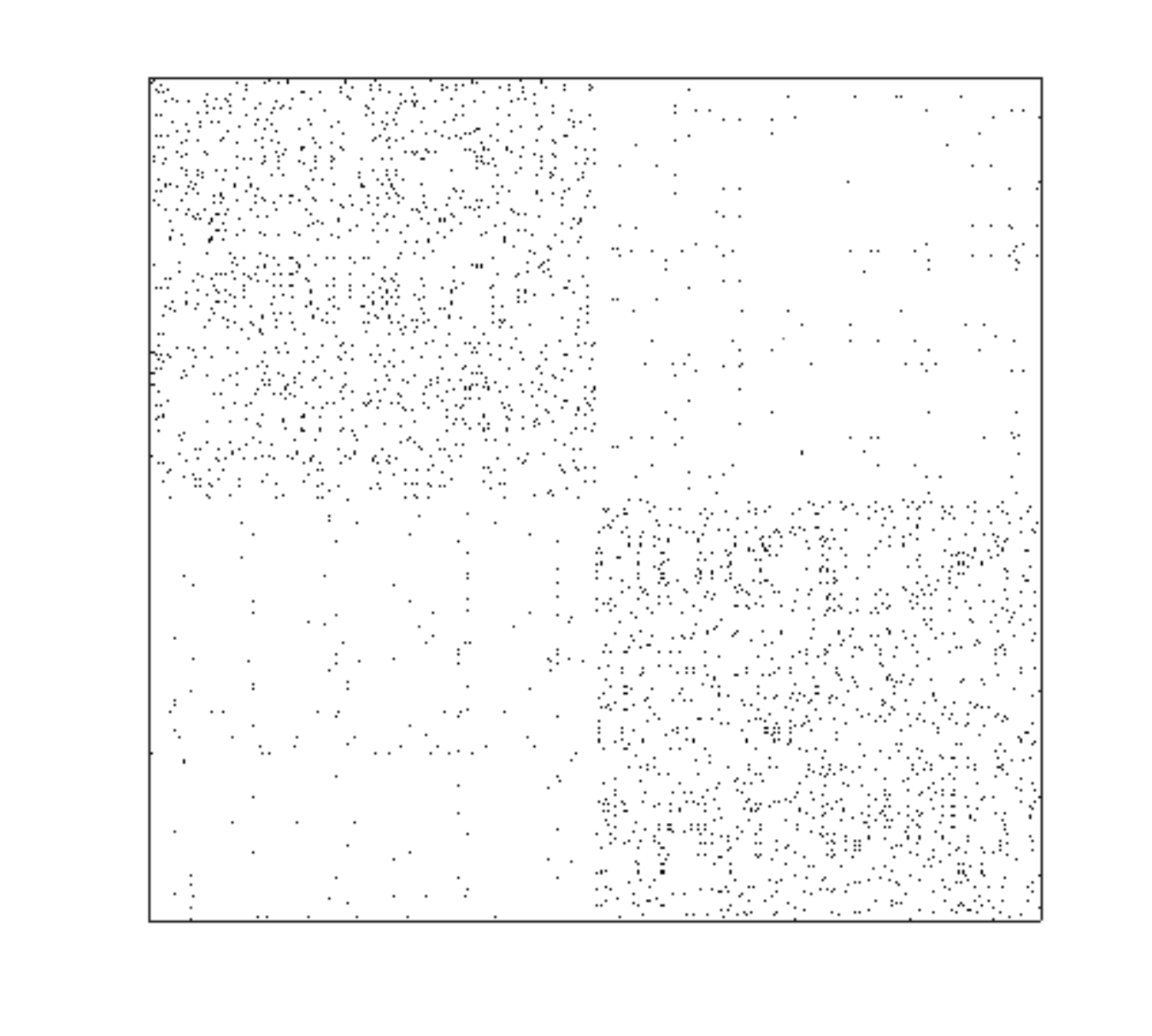}
}\label{s2}
\caption{Two-view synthetic dataset and the learned similarity matrix $S$ corresponding each individual input view.}
\label{toy}
\end{figure*}

\begin{table*}[t]
\centering
\caption{Clustering results and the learned view weights}\label{toy_table}
\begin{tabular}{c||c|c||c|c|c}  % {lccc} 表示各列元素对齐方式，left-l,right-r,center-c
\hline\hline
&\multicolumn{2}{|c||}{Single view learning} & \multicolumn{3}{|c}{CLR-IW iterative results} \\\cline{2-4}\hline
&View1 & View2   & $k=1$    & $k=2$   &$k=3$    \\\hline
ACC/NMI/Purity&0.950/0.732/0.950  &0.889/0.497/0.889   &0.976/0.838/0.976   &0.982/0.873/0.982   &\textbf{0.985}/\textbf{0.889}/\textbf{0.985}   \\\hline
View weights  &1/0  &0/1   &0.5/0.5  &0.515/0.485   &0.538/0.462   \\\hline\hline
\end{tabular}
\end{table*}
The first toy example is used to explain why we prefer multi-view learning rather than simply selecting a good view. We design a two-view synthetic dataset where each view is a 90 $\times$ 90 matrix with three 30 $\times$ 30 block matrices diagonally arranged. Without loss of generality, the data within each block denotes the affinity of two corresponding points in one cluster, while the data outside all of blocks denotes noise. Each element in all blocks is randomly generated in the range of 0 and 1, while the noise data is randomly generated in the range of 0 and e, where e is set as 0.6 in the 1st matrix, and 0.7 in the 2nd matrix. Following the complementary principle, in view 1 we increase the noise between the first and second block data to $e = 0.8$, and increase the noise between the second and third block data to $e = 1.0$. Then they are normalized to be that the sum of each row is 1. The original input graphs are shown in Figure \ref{toy1}. By performing CLR on each individual graph, their clustering performance are : ACC(Purity) of view 1 and 2 are 0.663/0.635, NMI of view 1 and 2 are 0.580/0.580. However, when CLR-IW is used to integrate these two complementary graphs, we recover the perfectly clean block diagonal matrix and the learned weights are 0.528/0.472, which indicates effectiveness of multi-view learning.

The second toy example works on a synthetic dataset given by \cite{kumar2011co} which contains two views that are showed as Fig. \ref{toy} (a) and (c) respectively. It can be observed that two-class samples are more discriminative in view 1 than they are in view 2. Thus, when each of them is applied to CLR, the similarity matrix $S$ learned by the former is more cleaner (See Fig. \ref{toy} (b) and (d). For better presentation, we normalize each similarity matrix by binaryzation in gray style). Table \ref{toy_table} shows the results of both single view clustering and iterative results using CLR-IW. It is seen that the "$k=3$" obtains the best clustering result, whose ACC, NMI, and Purity reach up to 0.985, 0.889, and 0.985. Moreover, each single view learning results and "$k=1$" being the baselines in this experiment, the generated CLR-IW has a noticeable improvement to them. Interestingly, since view 1 is stronger than view 2, the finally learned normalized weights are 0.538/0.462, which exactly agrees with our prior knowledge.

\subsection{Datasets}
The Datasets used in the following experiments are very popular in multi-view learning tasks, which are MSRC-v1 \cite{DBLP:conf/iccv/WinnJ05}, Caltech101 \cite{fei2007learning}, Handwritten numerals \cite{asuncion2007uci}, NUS-WIDE Animals \cite{chua2009nus} and MNIST \cite{lecun2010mnist}. The brief description of each dataset is introduced as follows:

1) \emph{MSRC-v1:} This collection is a scene recognition dataset which contains 240 images. Following \cite{grauman2006unsupervised}, we select 7 classes which are composed of tree, building, airplane, cow, face, car, bicycle, and each class has 30 images. We extracted five visual features for each image: 24 Color Moment, 576 HOG, 512 GIST, 256 LBP, and 254 CENTRIST.

2) \emph{Caltech101:} This dataset contains 8677 images which can be divided into 101 classes. We use two regular subsets Caltech101-7 and Caltech101-20 in our experiments. Six extracted features can be used, and they are 48 Gabor, 40 Wavelet Moments (WM), 254 CENTRIST, 1984 HOG, 512 GIST, 928 LBP.

3) \emph{Handwritten Numerals:} This dataset is about handwritten numerals (0-9) extracted from a collection of Dutch utility maps. There are 2000 patterns and 200 for each class. These digits are represented as six public features: 76 Fourier coefficients of the character shapes (FOU), 216 profile correlations (FAC), 64 Karhunen-love coefficients (KAR), 240 pixel averages in 2 $\times$ 3 windows (PIX), 47 Zernike moment (ZER) and morphological (MOR) features.

4) \emph{NUS-WIDE:} The dataset contains 269,648 images of 81 concepts. In our experiments, 12 categories about animal concept are selected and each contains 200 images. They are cat, cow, dog, elk, hawk, horse, lion, squirrel, tiger, whales, wolf, and zebra. Each image is represented by six type low-level features: 64 color histogram, 144 color correlogram, 73 edge direction histogram, 128 wavelet texture, 225 block-wise color moment and 500 bag of words based on SIFT descriptions.

5) \emph{MNIST:} The dataset of handwritten digits (0-9) from Yann LeCun's MNIST page has a test set of 10000 samples. There digits are described by three features: 30 isometric projection, 9 linear discriminant analysis and 30 neighborhood preserving embedding.

The statistics of these datasets are summarized in Table \ref{table_dataset}, where MNIST is only for graph-free methods, since graph-based clustering method is time-consuming when datasize is very large. For the space limitation, we use the abbreviations (MSRC, Cal-7, Cal-20, HN, and NUS) for the name of each dataset.

\begin{table}[t]
\centering
\caption{Statistics of four datasets}\label{table_dataset}
\begin{tabular}{c|c|c|c}  % {lccc} 表示各列元素对齐方式，left-l,right-r,center-c
\hline
Datasets & \# of data & \# of view & \# of cluster\\ \hline
MSRC-v1 & 210 & 5 &7 \\
Caltech101 &1474(2386)&6 & 7(20)\\
Handwritten Numerals & 2000 & 6 & 10\\
NUS-WIDE & 2400 & 6 & 12\\
MNIST & 10000 & 3 & 10\\ \hline
\end{tabular}
\end{table}

\subsection{Weight Learning Comparison}
\begin{table*}[tb]
\centering  % 表居中
\caption{Summaries of formulations which are generated by employing different weight learning approaches for CLR learner, where the target variable $S$ is constrained as it is in Eq. \eqref{eq3.2.1}, and $\alpha$ is constrained by $\mathcal{C}_{\alpha}$. }\label{compare_methods}
\begin{tabular}{l|c|c|c}
\hline
Methods & Objectives   & Hyper-parameter (Grid search)    & Referred work  \\ \hline
CLR-NR &\(\mathop {\min }\limits_{S,\alpha} \sum\limits_{v = 1}^M {{\alpha _v}\left\| {S - {A^{( v )}}} \right\|_F^2}  + {\gamma _1}\left\| \alpha  \right\|_2^2\) & \({\gamma _1} \ge 0\),  $[1, 5, 10, 50, 100, 500, 1000]$& \cite{cai2014feature,xu2016weighted,kumar2015unsupervised,karasuyama2013multiple}\\
CLR-ER &\(\mathop {\min }\limits_{S,\alpha } \sum\limits_{v = 1}^M {\big( {{\alpha _v}\left\| {S - {A^{( v )}}} \right\|_F^2 + {\gamma _2}{\alpha _v}\log {\alpha _v}} \big)} \) & \(\gamma_2 \ge 0\),   $[1,5,10,50,100,500,1000]$ & \cite{lange2005fusion,zhang2016weighted}\\
CLR-EF &\(\mathop {\min }\limits_{S,\alpha } \sum\limits_{v = 1}^M {\alpha _v^{{\gamma _3}}\left\| {S - {A^{\left( v \right)}}} \right\|_F^2} \) & \(\gamma_3 \ge 1\), $[1.5, 2.0, 2.5, 3.0, 3.5, 4.0]$ & \cite{xue2015gomes,tzortzis2012kernel,xu2016weighted,zhang2016weighted,tzortzis2010multiple} \\
CLR-IW &\(\mathop {\min }\limits_S \sum\limits_{v = 1}^M {\left\| {S - {A^{\left( v \right)}}} \right\|_F^p} \)  & \(0 < p \le 2\),   $[0.1,0.4,0.7,1.0,1.3,1.7]$&-\\
\hline
\end{tabular}
\end{table*}

\begin{figure*}[tb]
\centering
\subfigure[ACC]{
\includegraphics[width=0.31\textwidth]{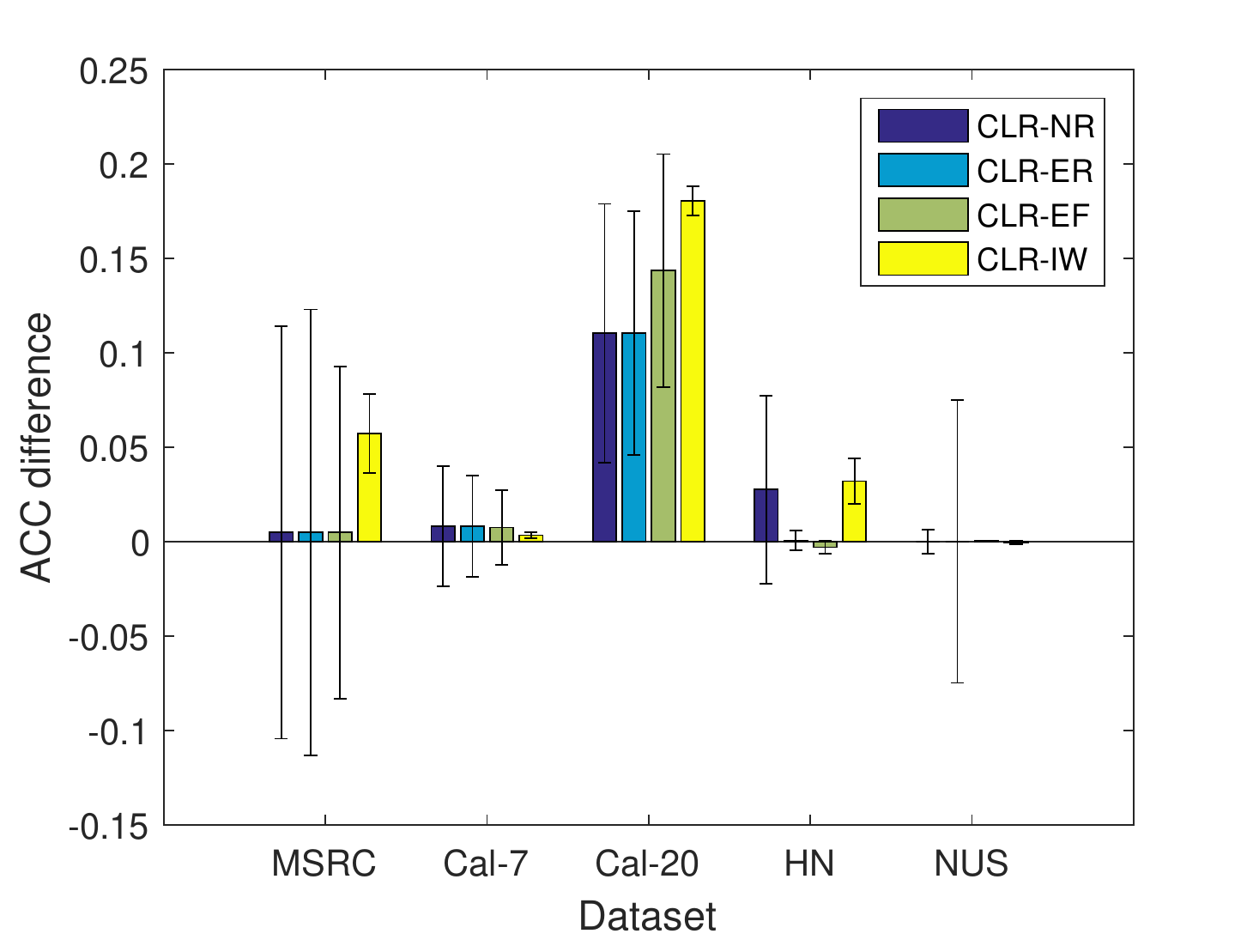}
}\label{acc}
\subfigure[NMI]{
\includegraphics[width=0.31\textwidth]{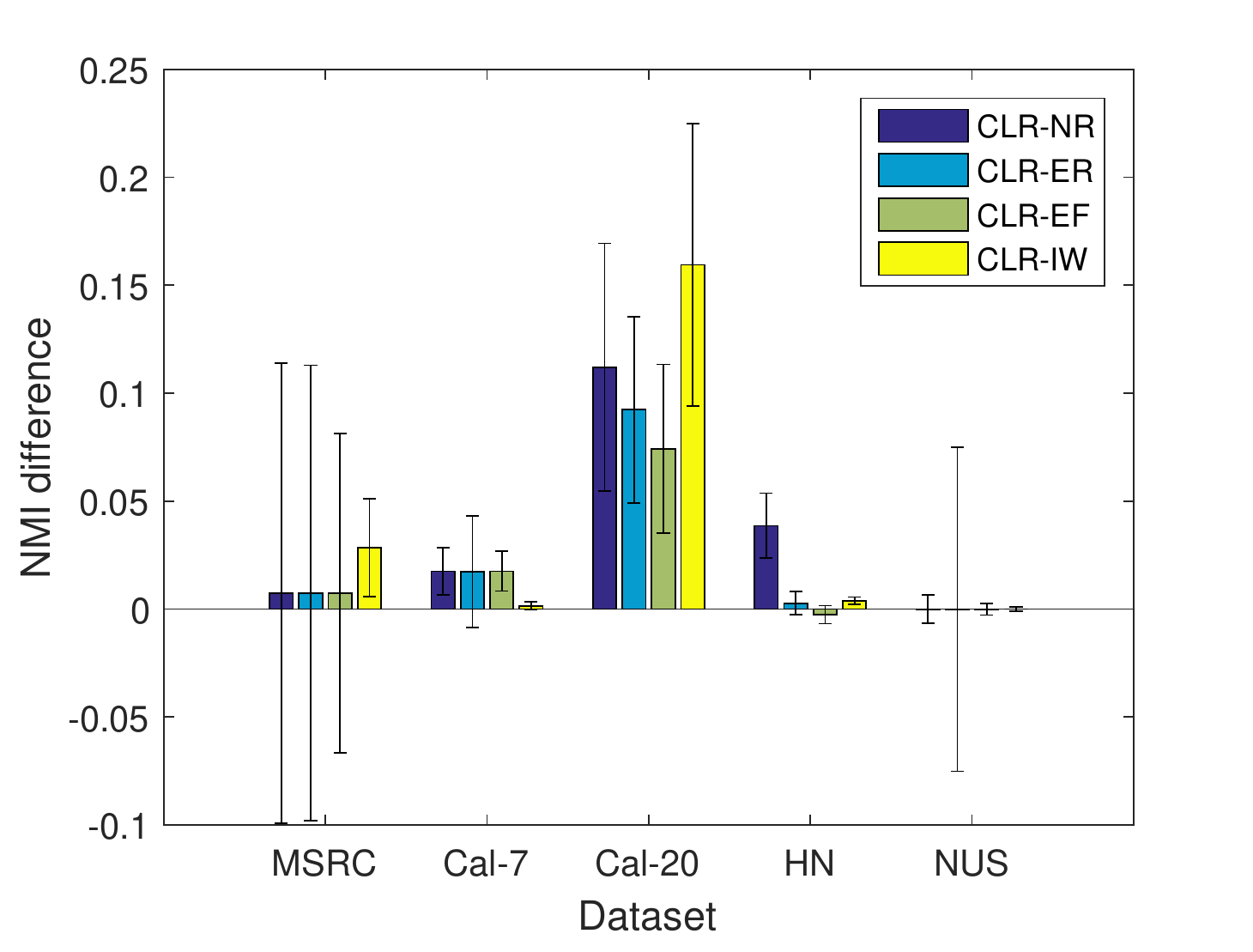}
}\label{nmi}
\subfigure[Purity]{
\includegraphics[width=0.31\textwidth]{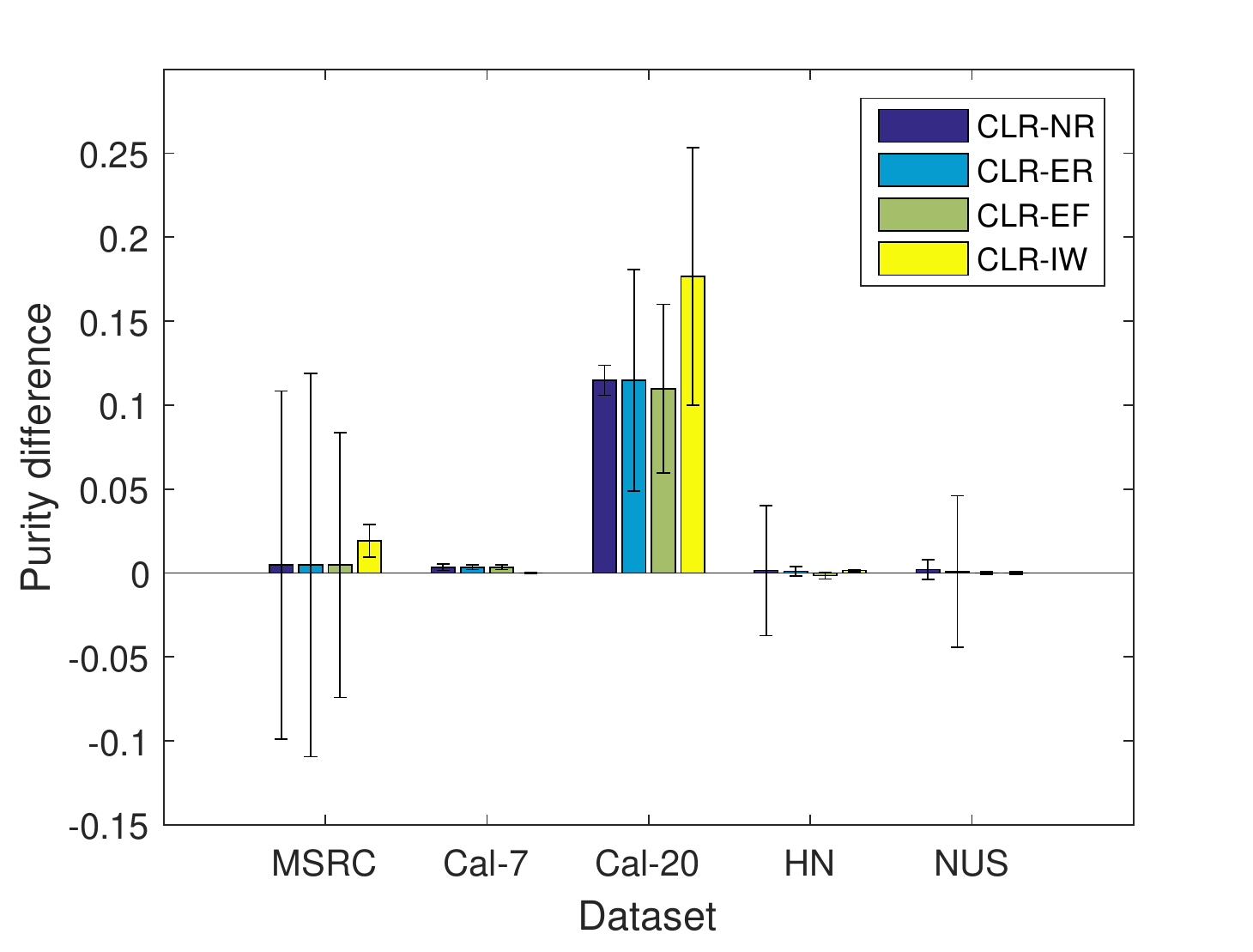}
}\label{pur}
\caption{Clustering performance of the generated methods listed in Table \ref{compare_methods} on various datasets.}
\label{multi_CLR_fig}
\end{figure*}

\begin{figure*}[tb]
\subfigure[CLR-NR]{
\includegraphics[width=0.22\textwidth]{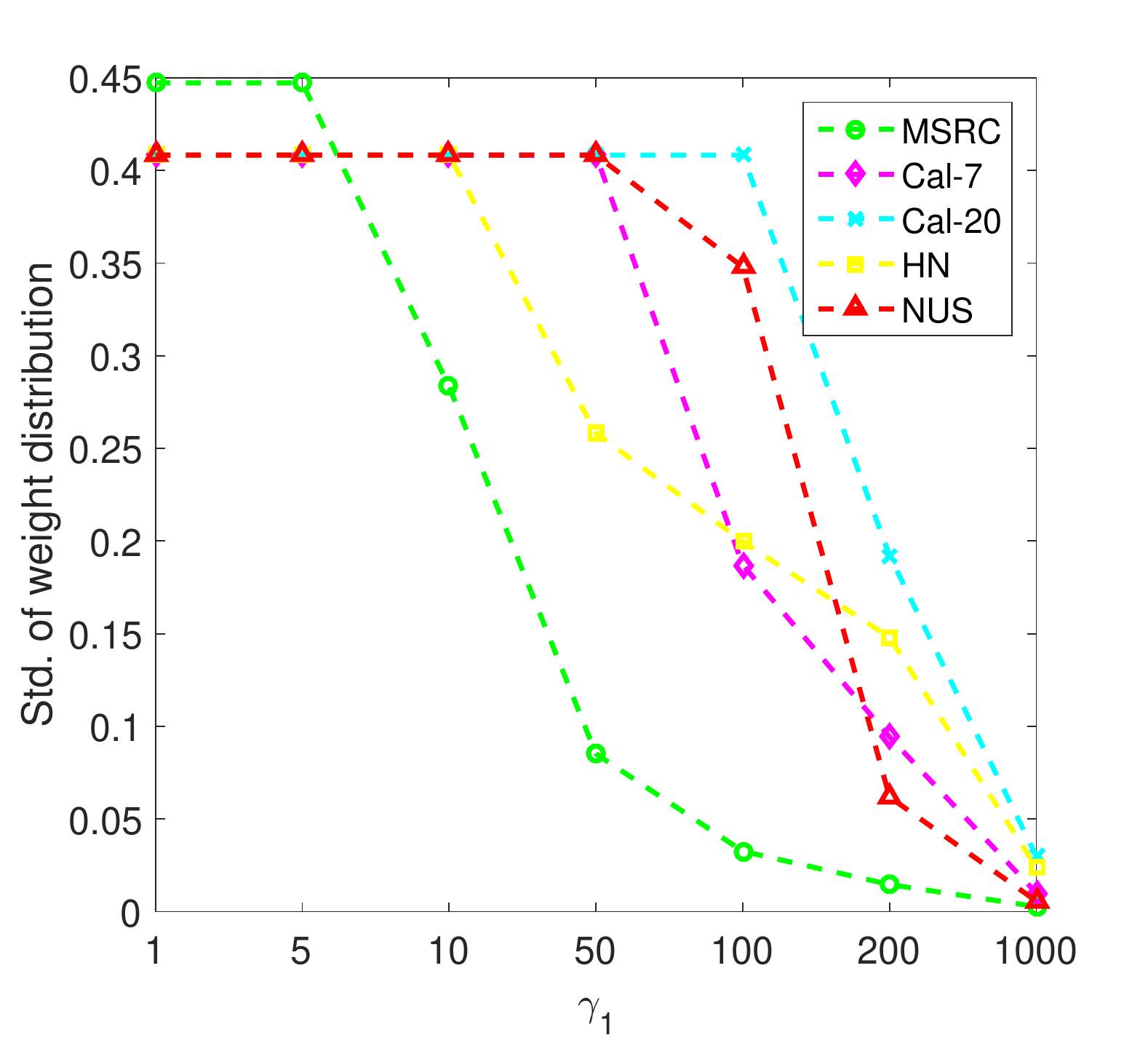}
}\label{CLR-NR}
\hfill
\subfigure[CLR-ER]{
\includegraphics[width=0.22\textwidth]{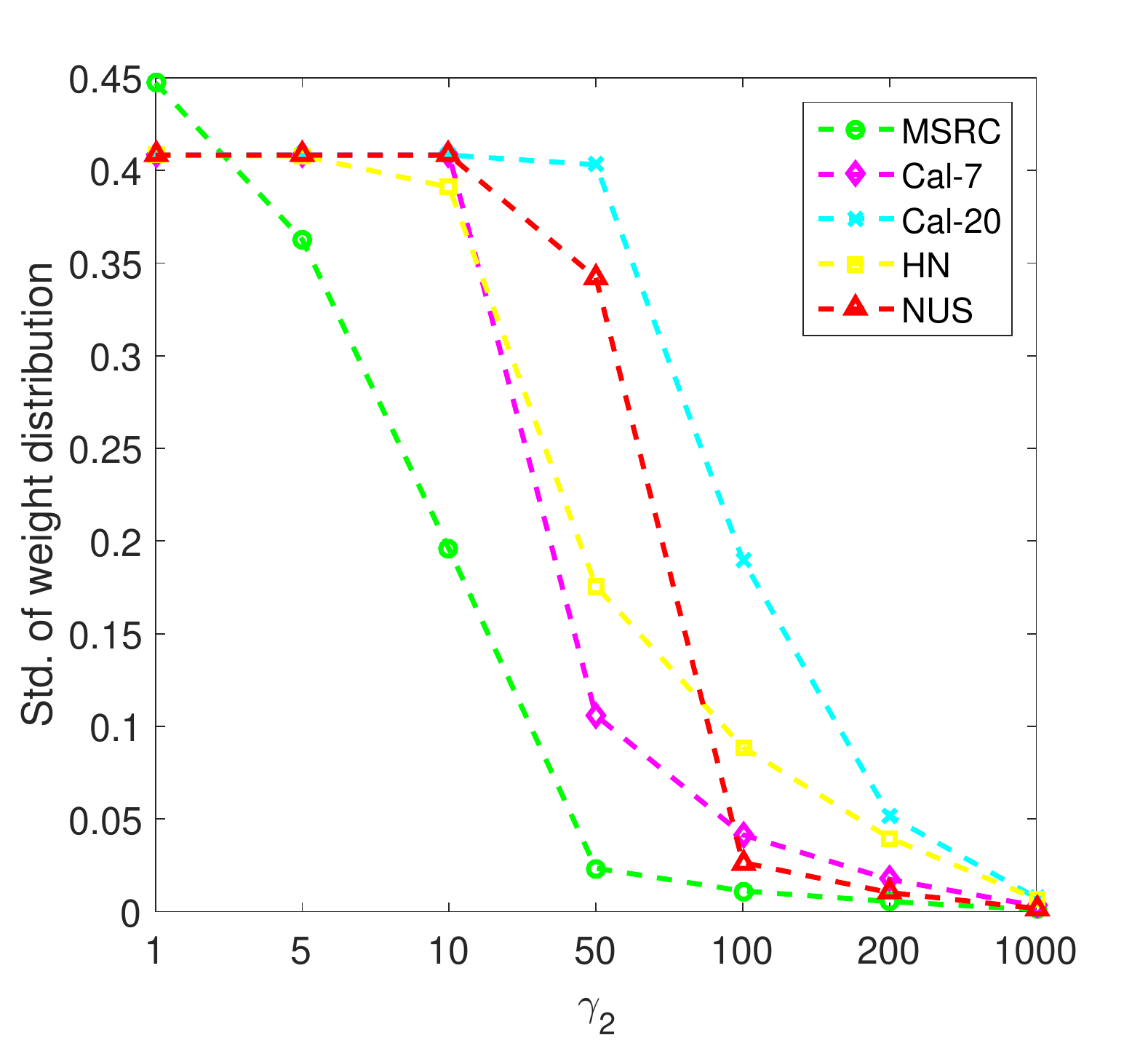}
}\label{CLR-ER}
\hfill
\subfigure[CLR-EF]{
\includegraphics[width=0.22\textwidth]{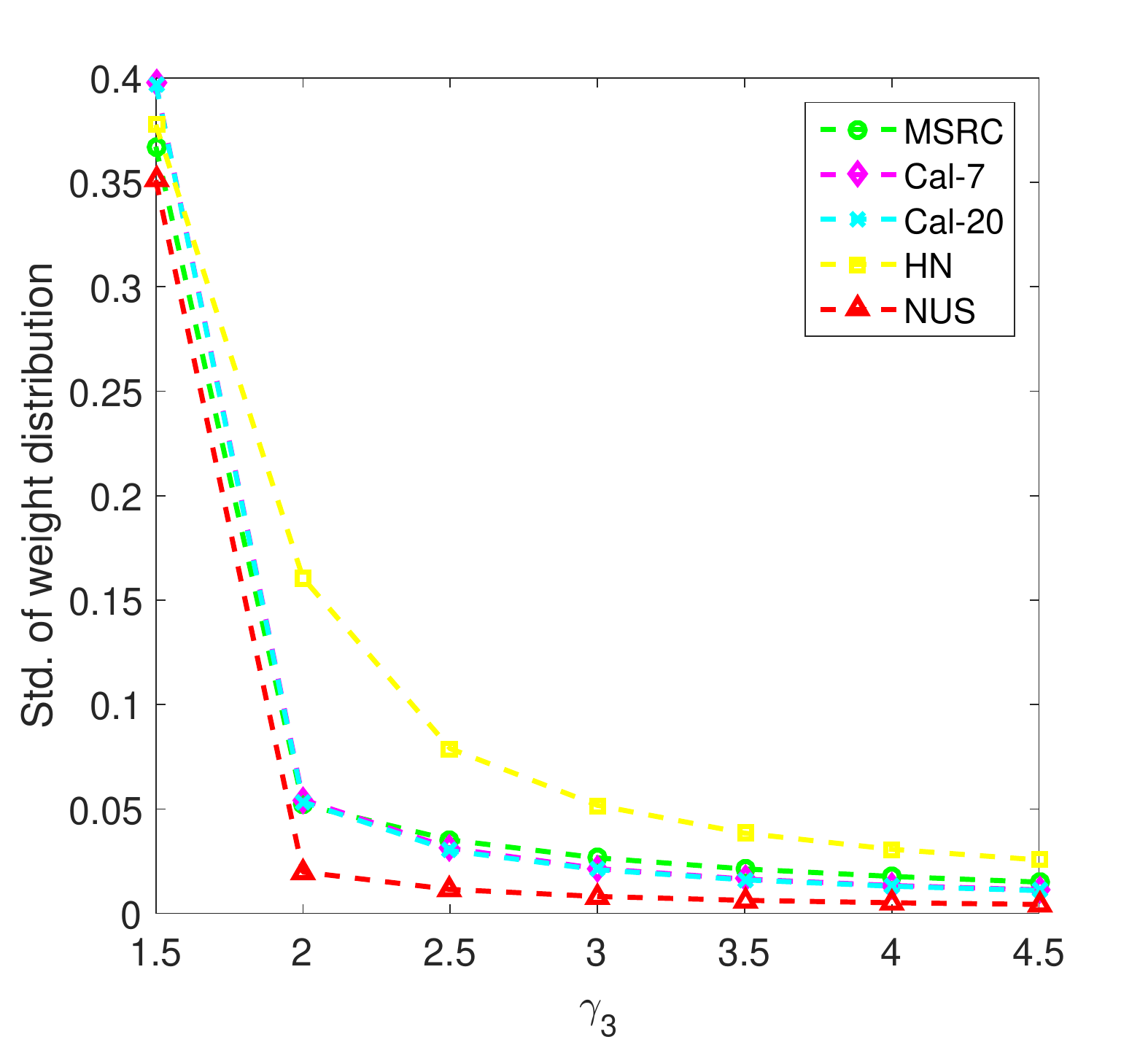}
}\label{CLR-EF}
\hfill
\subfigure[CLR-IW]{
\includegraphics[width=0.22\textwidth]{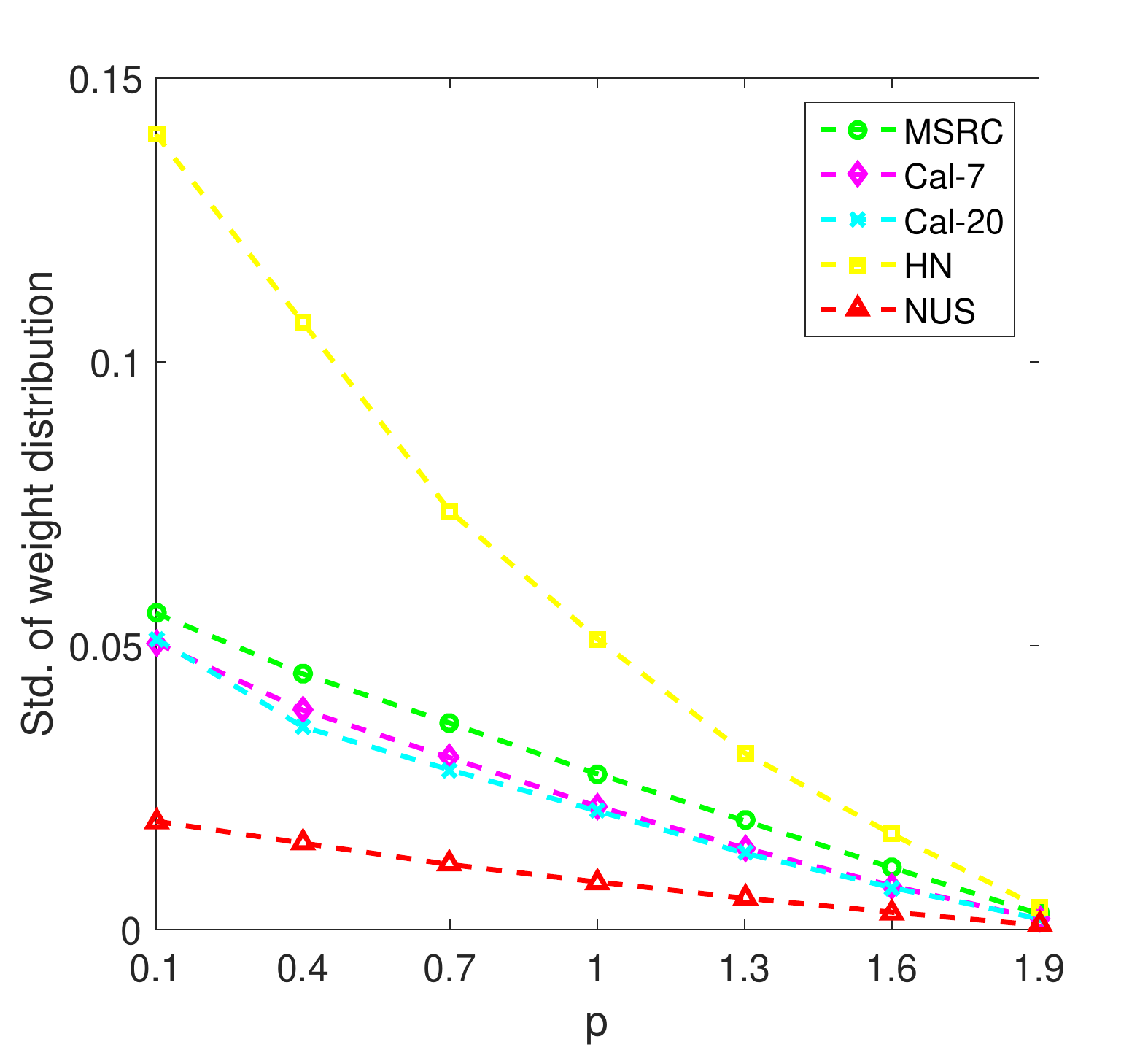}
}\label{CLR-IW}
\caption{Standard deviation of view weights for different methods on various datasets.}
\label{std_weight}
\end{figure*}

To quantitatively investigate the differences among all of aforementioned weight learning approaches, we do crosswise comparison by achieving each of them on CLR model rather than directly include any particular previous method. Their names, objectives, the corresponding parameters, and the referred work are showed in Table \ref{compare_methods}. As a convention in \cite{kumar2015unsupervised,xu2016weighted,zhang2016weighted,tzortzis2012kernel}, grid search is adopted here to try the different values of the hyper-parameter for each compared method. Although it is known that the wider and denser the grid is, the better clustering performance will be, it is very hard to set the proper range and step size of them in practical applications. In this paper, we empirically set them as in third column of Table \ref{compare_methods} after several small tests.

According to the analysis in Section \ref{sec:related works}, we easily come to the algorithm for each problem, which is ignorant in this part. It is obvious that each of them models a  non-convex problem, since the rank constraint is always supposed to be satisfied. Considering that in most cases every view contributes to the final clustering results, we initialize each algorithm with the equal weights. Fig. \ref{multi_CLR_fig} reports the clustering results on various datasets in terms of aforementioned three standard evaluation metrics respectively. Multi-view equal weights CLR is the potential baseline in this experiment, and the magnitude of a bar denotes how many improvements of each method makes. The standard deviation for each method is also presented by attaching the top of a bar.

At the first sight of Fig. \ref{multi_CLR_fig}, except NUS dataset, on which each method obtains the very close clustering results with equal weights version, it is observed that almost every method achieves better performance than the baseline, which indicates the weight learning does work for multi-clustering. More importantly, the best clustering results of CLR-IW most times are higher than others in compared schemes. It is because that the hyper-parameter for the intrinsic weight learning approach is searched in a interval with two-side boundary. On this event, grid search is often effective. Furthermore, from the standard deviation of each method, we know that the different values of hyper-parameter have an significant impact on the final clustering performance, and CLR-IW is robuster than any competing method.

Another point which we concern is the learned view weights. Standard deviation (Std) is used to describe the smoothness of the learned weights as in Fig. \ref{std_weight}. Generally, with the increase of the hyper-parameter in each compared method, Std is dropping sustainedly with a lower bound zero, which indicates the weights distributions are getting smoother. Particularly, when Std is zero, all of view weights become equal. In addition, we find that NR has the similar weight distributions with ER, which is apparently due the parallel type regularization form. As we analyze before, they both can learn the sparse weights, thus when the weight of regularization is small, it comes to very high Std. For EF, Std decreases dramatically when $\gamma_3$ varies form 1.5 to 2.
As the adopted grid search is following \cite{tzortzis2012kernel}, we do not subdivide this range any longer. Interestingly, when $\gamma_3 > 0$, EF is much similar with IW, i.e., they have the close values of Std and consistent order of datasets on each hyper-parameter point. The reason of this phenomenon can be simply explained as:

When Algorithm \ref{Alg0} converges, according to Eq. \eqref{eq4.4}, we know that the normalized weight \(\widetilde {\alpha_v} \) (like constraint $\mathcal{C}_{\alpha}$) can be represented as
$\widetilde {{\alpha _v}} = \frac{{\frac{p}{2}\Phi _v^{\frac{{p - 2}}{2}}}}{{\frac{p}{2}\sum\limits_{u = 1}^M {\Phi _u^{\frac{{p - 2}}{2}}} }} = \frac{1}{{\sum\limits_{u = 1}^M {{{\left( {\frac{{{\Phi _v}}}{{{\Phi _u}}}} \right)}^{\frac{2}{{p - 2}}}}} }}$,
which is identical with Eq. \eqref{eq2.8} when $ p = 2\gamma_3$.

Noting that the peak value in IW is not as large as other methods, which shows that IW actually prefers the smooth weight distribution.

%For CLR-IW, the sharpest weight distribution is not learned at the smallest $p$, which agrees with our analysis in Section \ref{sec:IWLA}. In addition, we find that t

\subsection{Extension to Other Clustering Tasks}
\begin{table*}[tb]
\centering  % 表居中
\caption{Summaries of formulations which are generated by employing intrinsic weight learning approach for NMF and SC clustering model, where $0<p\le2$. }\label{extention_two}
\begin{tabular}{lcc}  % {lccc} 表示各列元素对齐方式，left-l,right-r,center-c
\hline
Methods & Objectives   & Constriants     \\ \hline
SC-IW &\(\mathop {\min }\limits_G \sum\limits_{v = 1}^M {{{\left( {Tr\left( {{G^T}LG} \right)} \right)}^{\frac{p}{2}}}} \) & ${G^T}G = I$, or ${G^T}DG = I$   \\
NMF-IW &\(\mathop {\min }\limits_{F^{(v)},G} \sum\limits_{v = 1}^M {\left\| {{X^{{{(v)}}}}^T - G{F^{(v)}}^T} \right\|_F^p} \)  & \({G_{ic}} \in \left\{ {0,1} \right\},\sum\limits_{c = 1}^C {{G_{ic}}}  = 1,\forall i = 1,2,...,N\)\\
\hline
\end{tabular}
\end{table*}

Besides the recent clustering method CLR, the proposed intrinsic weight learning approach is easily extended to some other basic clustering approaches which produces more new multi-view clustering methods. Here we firstly introduce two base learners.
\begin{enumerate}[i]
\item Spectral Clustering (SC) method is a representative paradigm for nonlinear data clustering. Given the adjacent matrix $W \in \mathbb{R}^{N \times N}$ (The corresponding degree matrix and Laplacian matrix is $D$ and $L$), the clustering objective is
\[\mathop {\min }\limits_{G} Tr\left( {{G^T}LG} \right).\] If the constraint to $G$ is $G^TG = I$, it will come to Ratio Cut (RC) problem, while if the constraint is $G^TDG=I$, it becomes Normalized Cut (NC) problem.

\item Previous work \cite{ding2005nonnegative} proved that the G-orthogonal non-negative matrix factorization (NMF) is equivalent to relaxed $K$-means clustering, and the following objective which is formalized by NMF but embedding $K$-means is a popular clustering learner

\begin{equation*}
\begin{split}
&\mathop {\min }\limits_{F,G} \left\| {{X^T} - G{F^T}} \right\|_F^2 \\
s.t.\,{G_{ic}} \in & \left\{ {0,1} \right\}, \sum\limits_{c = 1}^C {{G_{ic}}}  = 1,\forall i = 1,2,...,N ,
\end{split}
\end{equation*}

where $X \in \mathbb{R}^{d \times N}$ is the input data matrix with $N$ samples and $d$-dimensional feature, $F \in \mathbb{R}^{d \times C}$ is the cluster centroid matrix, and $G \in \mathbb{R}^{N \times C}$ is the cluster assignment matrix and whose each row is the 1-of-$C$ coding scheme.
\end{enumerate}
By employing the proposed weight learning strategy to above two base learners, we obtain two multi-view clustering methods in Table \ref{extention_two}. The optimization procedure for SC-IW is quite simple while it is not direct for NMF-IW. So we show the detailed steps for NMF-IW in Appendix. The clustering results on various datasets are presented in Tables \ref{t5} and \ref{t6}, where we report the improvements of the best clustering performance relative to baselines. Since there are three metrics, we sum them up and find the largest one during the grid search.

Overall, we can see that with the use of intrinsic weight learning paradigm, the obtained clustering results apparently outperform baselines which does not take the discriminative view weights. Interestingly, it is observed that when intrinsic weight learning is applied in different base learners, the improvements sometimes discord. For instance, CLR-IW does not work well on NUS but SC-IW and NMF-IW do. This indicates that view weights are not absolute and actually much related to the specific model presentation capacity.

\begin{table}[tb]
\centering  % 表居中
\caption{Improvements of SC-IW with multi-view equal weights SC being the baseline. (\%) }
\label{t5}
\begin{tabular}{lc|c|c|c|c|c}  % {lccc} 表示各列元素对齐方式，left-l,right-r,center-c
\hline
&        & MSRC   & Cal-7 & Cal-20  & HN   & NUS     \\ \hline
\multirow{3}{*} {RC-IW}&
ACC     &  +0.48 & +8.35 & +8.22 & +11.00 & -1.30\\
&NMI     &  +1.40 & +5.69 & +3.82 & +6.42 & -1.37\\
&Purity  &  +4.29 & -9.32 & -3.18 & +17.55 & +14.17\\
\hline
\multirow{3}{*} {NC-IW}&
ACC     &  +0.47 & +6.31 & -1.00 & +2.35 & -1.75\\
&NMI     &  +0.15 & +1.77 & +2.04 & +2.61 & -1.16\\
&Purity  &  +4.28 & -6.55 & +9.30 & +9.85 & +22.08\\
\hline
\end{tabular}
\end{table}

\begin{table}[tb]
\centering  % 表居中
\caption{Improvements of NMF-IW with multi-view equal weights NMF being the baseline. (\%) }
\label{t6}
\begin{tabular}{l|c|c|c|c|c|c}  % {lccc} 表示各列元素对齐方式，left-l,right-r,center-c
\hline
        & MSRC   & Cal-7 & Cal-20  & HN   & NUS & MNIST     \\ \hline
ACC     &  +9.00 & +0.80 & +3.23 & +16.85 & +6.00 & +7.53\\
NMI     &  +4.90 & +0.80 & -0.58 & +12.10 & +3.48 & +7.32\\
Purity  &  +8.60 & +0.20 & -1.34 & +16.85 & +6.34 & +7.24\\
\hline
\end{tabular}
\end{table}

\section{Conclusion}
\label{sec:conclusion}
In this paper, we present a new weight learning strategy named Intrinsic Weight Learning Approach for multi-view clustering task. By comparing with several classical weight learning approaches both in theory and experiments, we conclude that the proposed intrinsic weight learning is robuster to hyper-parameter and easy to obtain the better results. Moreover, we show that the proposed weight learning approach can be naturally used in other basic clustering learners. Extensive experiments have shown that this weight learning approach significantly improves the clustering performance and practical to use. In the future work, like \cite{nie2016parameter} we will consider formulating a general framework which can also work in semi-supervised context.

%\appendices
%\section{Proof of the optimization procedure of solving NMF-IW problem}
\appendix[The optimization procedure of solving NMF-IW problem]
According to Algorithm \ref{Alg0}, we know that the key step to address the NMF-IW problem is to solve the following subproblem
\begin{equation}
\label{eq:a1}
\mathop {\min }\limits_{\alpha ,G,{F^{\left( v \right)}}} \sum\limits_{v = 1}^M {{\alpha _v}\left\| {\mathop {{X^{\left( v \right)}}}\nolimits^T  - G\mathop {{F^{\left( v \right)}}}\nolimits^T } \right\|} _F^2,
\end{equation}
where $\alpha$ is fixed.
This problem can be solved by alternatively update $F^{(v)}$ and $G$ in each iteration. When $G$ is fixed, taking the derivative of Eq. \eqref{eq:a1} w.r.t $F^{(v)}$ and setting the derivatives to zeros, for each $1 \le v \le M$, we have
\begin{equation}
\begin{split}
&{\alpha _v}{\left( {\mathop {{X^{\left( v \right)}}}\nolimits^T  - G} \right)^T}\left( { - G} \right) = {\bf{0}}\\
 \Rightarrow & {F^{\left( v \right)}} = XG{\left( {{G^T}G} \right)^{ - 1}}.
\end{split}
\end{equation}
When $G$ is fixed, Eq. \eqref{eq:a1} can be split into $n$ smaller subproblems, each of which can be written as ($1\le i \le N$)
\begin{equation}
\label{eq:a3}
\mathop {\min }\limits_{{g_i}} \sum\limits_{v = 1}^M {\left\| {{x^i} - {g_i}\mathop {{F^{\left( v \right)}}}\nolimits^T } \right\|_2^2} \quad {g_i} \in \left\{ {0,1} \right\},\sum\limits_{c = 1}^C {{g_{ic}}}  = 1.
\end{equation}
Seeing that $g_i$ satisfied $1$-$of$-$C$ coding scheme, there are $C$ candidates to be the solution of Eq. \eqref{eq:a3}, each of which is the $c$-th row of matrix ${I_C} = \left[ {{{\bf{e}}_1};{{\bf{e}}_2};...;{{\bf{e}}_C}} \right]$. This problem can be effectively solved by exhaustively searching strategy if $C$ is not very large.
%\section{Proof of the relation between EF and IW}
%When Algorithm \ref{Alg0} converges, according to Eq. \eqref{eq4.4}, we know that the normalized weight \(\widetilde {\alpha_v} \) can be represented as
%\begin{equation}
%\widetilde {{\alpha _v}} = \frac{{\frac{p}{2}\Phi _v^{\frac{{p - 2}}{2}}}}{{\frac{p}{2}\sum\limits_{u = 1}^M {\Phi _u^{\frac{{p - 2}}{2}}} }} = \frac{1}{{\sum\limits_{u = 1}^M {{{\left( {\frac{{{\Phi _v}}}{{{\Phi _u}}}} \right)}^{\frac{2}{{p - 2}}}}} }}.
%\end{equation}
%\section{Proof of the First Zonklar Equation}
%Appendix one text goes here.

% you can choose not to have a title for an appendix
% if you want by leaving the argument blank%\section{}
%Appendix two text goes here.

% use section* for acknowledgment
\ifCLASSOPTIONcompsoc
  % The Computer Society usually uses the plural form
 % \section*{Acknowledgments}
\else
  % regular IEEE prefers the singular form
 % \section*{Acknowledgment}
\fi

%The authors would like to thank...

% Can use something like this to put references on a page
% by themselves when using endfloat and the captionsoff option.
\ifCLASSOPTIONcaptionsoff
  \newpage
\fi

% trigger a \newpage just before the given reference
% number - used to balance the columns on the last page
% adjust value as needed - may need to be readjusted if
% the document is modified later
%\IEEEtriggeratref{8}
% The "triggered" command can be changed if desired:
%\IEEEtriggercmd{\enlargethispage{-5in}}

% references section

% can use a bibliography generated by BibTeX as a .bbl file
% BibTeX documentation can be easily obtained at:
% http://www.ctan.org/tex-archive/biblio/bibtex/contrib/doc/
% The IEEEtran BibTeX style support page is at:
% http://www.michaelshell.org/tex/ieeetran/bibtex/
%\footnotesize
\bibliographystyle{IEEEtran}
% argument is your BibTeX string definitions and bibliography database(s)

\bibliography{WMC}
\end{document}